\documentclass[10pt,twocolumn,letterpaper]{article}

\usepackage{cvpr}              

\usepackage{lipsum}
\usepackage{xspace}
\usepackage{graphicx}

\allowdisplaybreaks

\definecolor{cvprblue}{rgb}{0.21,0.49,0.74}
\usepackage[pagebackref,breaklinks,colorlinks,allcolors=cvprblue]{hyperref}

\usepackage{xcolor}
\usepackage{xspace}
\usepackage{colortbl}

\definecolor{colorTrd}{rgb}{0.95, 0.95, 0.75}
\definecolor{colorSnd}{rgb}{1, 0.85, 0.7}
\definecolor{colorFst}{rgb}{1, 0.7, 0.7}

\def\paperID{4713} 
\def\confName{CVPR}
\def\confYear{2026}

\def\method{PhyGaP\xspace} 
 
\def\polar{PolarDR\xspace} 
\def\subenvmap{GridMap\xspace} 

\title{\method: Physically-Grounded Gaussians with Polarization Cues}

\author{Jiale Wu\textsuperscript{1,2}\quad Xiaoyang Bai\textsuperscript{1,*}\quad Zongqi He\textsuperscript{1}\quad Weiwei Xu\textsuperscript{2}\quad Yifan Peng\textsuperscript{1,*}\\
{\small \textsuperscript{1}The University of Hong Kong\quad \textsuperscript{2}Zhejiang University}\\
{\tt\small jialewu2022@zju.edu.cn\quad xybai@hku.hk\quad zongqi\_he@connect.hku.hk}\\
{\tt\footnotesize xww@cad.zju.edu.cn\quad evanpeng@hku.hk}}

\begin{document}

\twocolumn[{%
\renewcommand\twocolumn[1][]{#1}%
\maketitle
\centering
\vspace{-0.6cm}
\includegraphics[width=\linewidth]{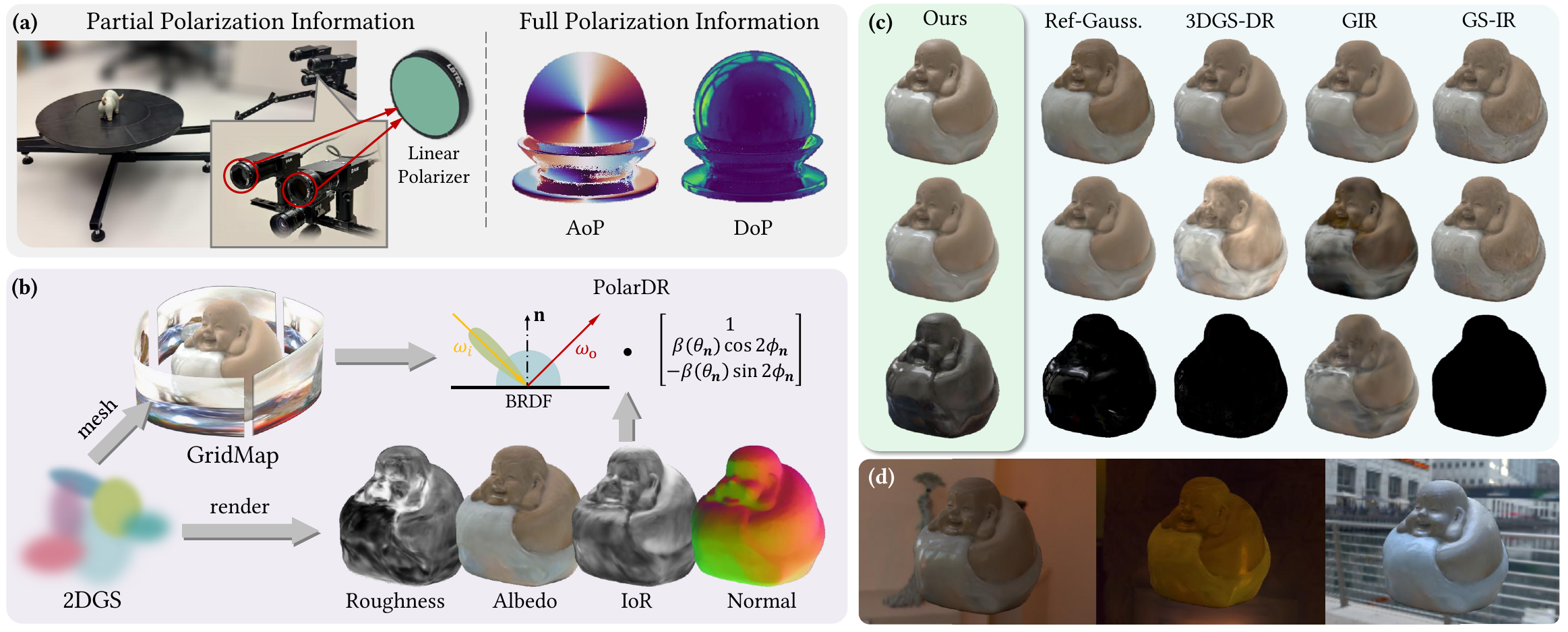}
\vspace{-19pt}
\captionof{figure}{We propose \textbf{\method}, a physically-grounded 3DGS method that (a) takes full or partial polarization information as input, (b) accurately reconstructs the shape and physical attributes of glossy object, (c) achieves decomposed rendering of object appearance (\textit{top}), diffuse reflection (\textit{mid}), and specular reflection (\textit{bottom}), as well as (d) enables robust and realistic relighting with natural reflection. Results in this visualization are from our captured \textit{buddha} scene.}\
\vspace{-2pt}
\label{fig:teaser}
}]

\footnotetext[1]{Corresponding authors.}
\begin{abstract}
Recent advances in 3D Gaussian Splatting (3DGS) have demonstrated great success in modeling reflective 3D objects and their interaction with the environment via \textbf{deferred rendering (DR)}. However, existing methods often struggle with correctly reconstructing physical attributes such as albedo and reflectance, and therefore they do not support high-fidelity relighting. Observing that this limitation stems from the lack of \textbf{shape and material} information in RGB images, we present \method, a physically-grounded 3DGS method that leverages polarization cues to facilitate precise reflection decomposition and visually consistent relighting of reconstructed objects. Specifically, we design a polarimetric deferred rendering (\polar) process to model polarization by reflection, and a self-occlusion-aware environment map building technique (\subenvmap) to resolve indirect lighting of non-convex objects. We validate on multiple synthetic and real-world scenes, including those featuring only partial polarization cues, that \method not only excels in reconstructing the appearance and surface normal of reflective 3D objects ($\sim$2~dB in PSNR and 45.7\% in Cosine Distance better than existing RGB-based methods on average), but also achieves state-of-the-art inverse rendering and relighting capability. Our code will be released soon.
\end{abstract}
\vspace{-12pt}
\section{Introduction}
\label{sec:intro}

Modeling reflective objects ranks among the most challenging task in 3D object reconstruction. For 3DGS~\cite{kerbl3Dgaussians} variants particularly, the lack of explicit geometry representation and the inability of the splatting pipeline to simulate secondary light transport limit their capability of representing glossy surfaces~\cite{fei20243d, bao20253d}. Only recently have several works emerged to tackle this bottleneck. To regularize surface smoothness, 2DGS~\cite{Huang2DGS2024} and Gaussian Surfel~\cite{dai2024high} equip Gaussian primitives with explicit normal attributes, while GaussTR~\cite{jiang2025gausstr} and Feat2GS~\cite{chen2025feat2gs} use foundation models such as Metric3D V2~\cite{hu2024metric3d} to provide surface normal supervision. On the other hand, the introduction of \textbf{deferred rendering (DR)} into GS training~\cite{Liang-gsir, ye2024gsdr} enables the modeling of reflection. Combining both branches together, state-of-the-art methods such as Ref-Gaussian~\cite{yao2025refGS} are able to accurately synthesize novel views of reflective objects.

However, objects reconstructed by those pipelines are prone to color shifts, unrealistic shading, or even surface discontinuities when placed under a different lighting condition, indicating incorrect decomposition between object albedo (inherent color) and reflected light (diffuse and specular) (Fig.~\ref{fig:teaser}(c)). This is because ray tracing-based DR relies on precise estimation of surface normal, reflectance, and roughness, which ordinary RGB images do not encode. Inspired by prior works that leverage \emph{polarization information} to learn those physical attributes~\cite{dave2022pandora, han2024nersp}, this work presents \textbf{\method}, a physically-grounded GS pipeline using polarization cues to enable \emph{reliable reflection decomposition} and supports \emph{coherent relighting} for glossy objects.

The \method framework aims to bridge the gap between Gaussian representations and the physical world via accurate reflection modeling. Specifically, we jointly model Gaussian appearance, shape, index of refraction (IoR), and roughness, as well as a learnable environment map. We further derive a polarimetric deferred rendering (\textbf{\polar}) process that computes pixel-wise Stokes values from this physically-grounded scene representation using the polarimetric BRDF (pBRDF) model, thus enabling direct scene optimization with either full or partial polarization information. Additionally, we present a self-occlusion-aware environment map building technique (\textbf{\subenvmap}) that addresses complex shading of non-convex objects without learning scene-specific indirect lighting, an improvement crucial for robustly relighting objects with arbitrary shapes.

We validate \method on both synthetic and real-world datasets of multiview polarization images, and evaluate its performance across different aspects: novel view synthesis (NVS) quality, normal accuracy, reflection decomposition, and relighting. Experimental results validate that its reconstruction of object appearance and normal is comparable to state-of-the-art reflection-aware methods, regardless of the availability of polarization cues, while it better estimates the albedo, diffuse, and specular reflection components of target scenes, demonstrating superior decomposition capability and leading to more realistic relighting results.

In summary, this work's contributions are threefold:
\begin{itemize}
    \item Our \method pipeline implements polarization-based GS optimization with a physically-grounded \polar process, facilitating accurate normal reconstruction and precise reflection decomposition of glossy objects.
    \item Our \subenvmap technique resolves indirect lighting and complex inter-reflection without learning and querying scene-specific parameters, expanding the relighting capacity of \method to non-convex objects.
    \item We validate our claim experimentally on a wide range of synthetic and real-world data. We also demonstrate the effectiveness of \method when only partial polarization information is available with real-world captures.
\end{itemize}

\section{Related Work}
\label{sec:rel}

\paragraph{Reflective Object Reconstruction and Inverse Rendering.}
As a long-standing topic in the graphics and vision communities, existing 3D reconstruction methods are mostly designed for opaque surfaces with the Lambertian reflectance model~\cite{lambert1760photometria,cook1982reflectance}. To model view-dependent visual features, Neural Radiance Field (NeRF)~\cite{mildenhall2021nerf} takes viewing direction as an argument of its scene function, while 3D Gaussian Splatting (3DGS)~\cite{kerbl3Dgaussians} leverages spherical harmonics (SH) for color representation. However, these tricks are not physically-grounded and thus still struggle with specular reflection~\cite{tosi2024nerfs}.

Attempts to extend 3D reconstruction methods to reflective objects started early in the evolution of NeRF. Works like Mirror-NeRF~\cite{zeng2023mirror}, NeRFReN~\cite{guo2022nerfren}, MS-NeRF~\cite{yin2023multi}, and NeRF-MD~\cite{van2025nerfs} focus exclusively on mirror and metal surfaces. Ref-NeRF~\cite{verbin2024ref} models view-dependent appearances with MLPs which requires a large number of parameters. Spec-NeRF~\cite{ma2024specnerf}, NDE~\cite{wu2024neural}, NeRF-Casting~\cite{verbin2024nerf}, and NeRO~\cite{liu2023nero} incorporate physics-based rendering with neural fields to achieve realistic reflection. Yet all these methods suffer from the problem of long runtime and heavy computation that all NeRF variants share.

The mainstream solution to modeling reflection with 3DGS is through \emph{deferred rendering}, which firstly rasterizes per-view physical attribute maps (albedo, roughness, metallic, etc.) through $\alpha$-blending, and then computes reflection from evaluating the rendering equation. To obtain accurate and consistent surface normals to estimate incident light direction, GShader~\cite{jiang2024gaussianshader}, GS-IR~\cite{Liang-gsir}, 3DGS-DR~\cite{ye2024gsdr}, and GIR~\cite{shi2025gir} directly use the shortest principal axis of each Gaussian; Relightable 3DGS (R3DG)~\cite{gao2024r3dg} instead designates normal as learnable attributes; and Ref-Gaussian~\cite{yao2025refGS} bases itself on 2DGS~\cite{Huang2DGS2024}. Despite their effectiveness, many of these methods do not correctly decompose albedo from reflection, and therefore can not support object relighting. We attribute this limitation to insufficient observability of per-pixel reflection state in RGB images alone, a gap we seek to bridge by leveraging polarization cues.

\begin{figure*}[t] 
\centering
\includegraphics[width=0.99\textwidth]{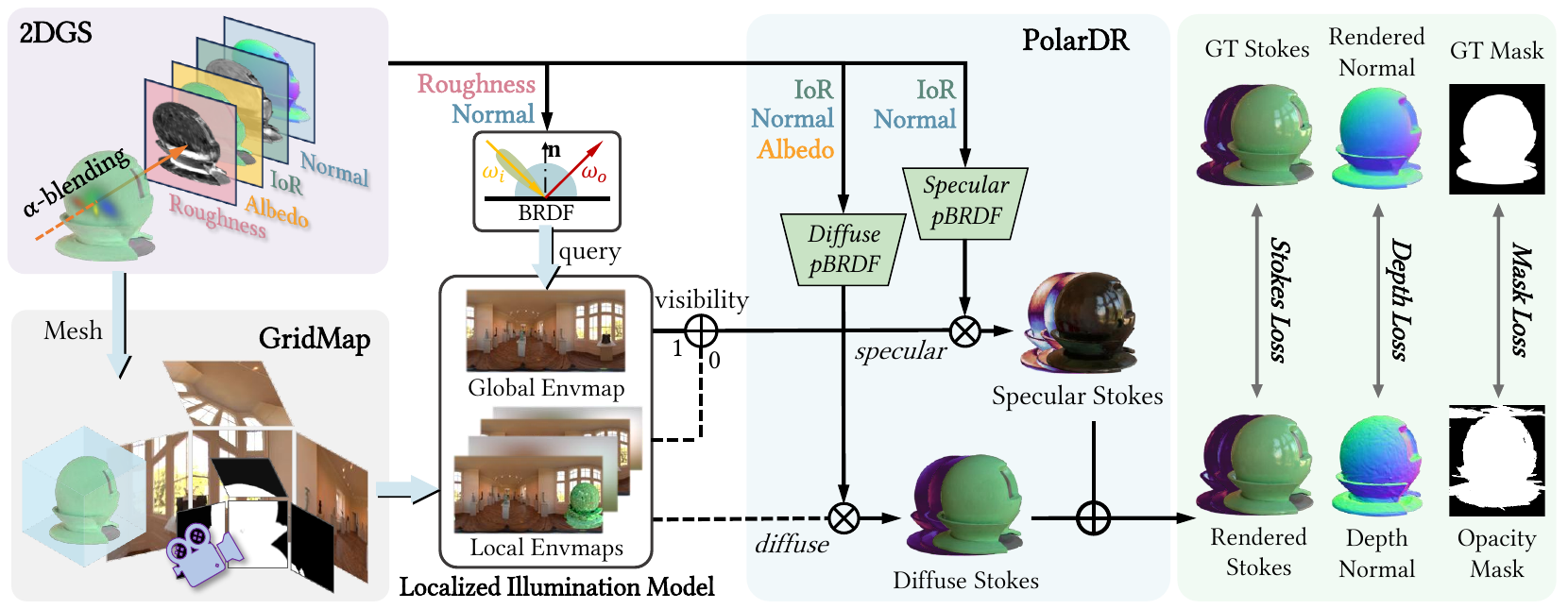}\vspace{-6pt}
\caption{Overview of the \method pipeline. We represent physically-grounded attributes such as roughness, albedo, IoR and surface normal with 2DGS, and render them into Stokes values via the \polar process. Furthermore, we design the \subenvmap technique to tackle self-occlusion of nonconvex objects. By utilizing polarization cues, we achieve \textit{accurate}, \textit{explicit} and \textit{disentangled} representation of object albedo, diffuse reflection and specular reflection in \method.}
\label{fig:pipeline}
\vspace{-9pt}
\end{figure*}

\vspace{-9pt}
\paragraph{Polarization-Based Vision and 3D Reconstruction.}
Polarization describes the orientation in which light wave oscillates. Since circular polarization is rarely observed in nature, this work is only concerned with \emph{linear polarization}. Linear polarization is known to encode rich cues on surface geometry and material, and therefore has been widely applied to estimate shape (SfP)~\cite{ba2020deep, deschaintre2021deep, tian2023dps, lei2022shape} and classify material~\cite{lin2022image, dong2024exploiting}. Its interaction with reflection and refraction further enables applications such as specular reflection removal~\cite{wen2021polarization}, dehazing~\cite{zhou2021learning}, underwater localization~\cite{bai2023polarization}, and time-of-flight imaging~\cite{baek2022all}.

Stemming from existing works on polarization-based multi-view stereo~\cite{fukao2021polarimetric, zhao2022polarimetric, cui2017polarimetric}, SLAM~\cite{yang2018polarimetric}, and depth-from-polarization~\cite{fukao2021polarimetric, blanchon2021p2d}, polarization-based NeRF has recently demonstrated state-of-the-art accuracy in modeling glossy objects. The pioneering PANDORA~\cite{dave2022pandora} learns to disentangle diffuse and specular reflection, and renders pixel-wise Stokes vectors via pSVBRDF~\cite{baek2018simultaneous}. NeRSP~\cite{han2024nersp} further utilizes polarization cues to achieve better surface reconstruction. Given the practical challenges of acquiring polarization images, Wu et al.~\cite{wu2025glossy} explore shape reconstruction from monocular linearly polarized inputs. 
The concurrent work PolGS~\cite{han2025polgs} is, to the best of our knowledge, the only attempt to fuse RGB and polarization modalities within 3DGS. However, none of these works supports relighting, which is enabled by our proposed \method.


\section{Preliminaries: \textit{Ref-Gaussian}}
\label{sec:prelim}

Our framework is built upon \textit{Ref-Gaussian}~\cite{yao2025refGS}, which leverages 2D Gaussian Splatting (2DGS)~\cite{Huang2DGS2024} for accurate geometry representation and deferred rendering (DR) to model reflection. We here give a brief overview of its pipeline before diving into \method itself.


\subsection{2D Gaussian Splatting} 

2DGS differs from 3DGS in that each Gaussian primitive is two-dimensional, defined by mean position $\textbf{p}$, two tangential vectors $\mathbf{t}_u$ and $\mathbf{t}_v$, and scaling factors $(s_u, s_v)$. The mapping from coordinates $(u,v)$ on each Gaussian's local tangent plane to the 3D world frame is thus given by:
\begin{align}
    P(u, v) &= \mathbf{p} + s_u\mathbf{t_u}\cdot u + s_v\mathbf{t_v}\cdot v  \notag\\
    &= \mathbf{H}\cdot\begin{bmatrix}
        u&v&1&1
    \end{bmatrix}^T,\\
    \text{where }\mathbf{H} &= \begin{bmatrix}
        s_u\mathbf{t}_u & s_v\mathbf{t_v} & 0 & \mathbf{p}\\
        0&0&0&1
    \end{bmatrix} \notag.
\end{align}

Then, the intersection between the camera ray passing through a given pixel $(x, y)$ and the tangent plane can be calculated by solving the equation~\cite{EWAvolumesplatting}:
\begin{equation}
    \begin{bmatrix}
        xz & yz & z & 1
    \end{bmatrix}^T = \mathbf{W}\mathbf{H}\cdot\begin{bmatrix}
        u&v&1&1
    \end{bmatrix}^T,
\end{equation}
where $\mathbf{W}$ is the world-to-camera transformation matrix. After $u$, $v$, and $z$ have been determined, we may then perform standard $\alpha$-blending, ordered by $z$ from front to back, where the influence of each Gaussian is defined as:
\begin{equation}
    \mathcal{G}(u, v) = \exp\left(-\frac{u^2+v^2}{2}\right).
\end{equation}

Notably, the normal of each Gaussian is easily computed as $\mathbf{n} = \mathbf{t}_u \times \mathbf{t}_v$, and a normal map can be subsequently rendered for each viewpoint using $\alpha$-blending.


\subsection{Deferred Rendering (DR)}

\emph{Ref-Gaussian} defines the following physical attributes to each Gaussian: albedo $\boldsymbol{\lambda}$, metallic $m$, and roughness $r$. Instead of directly splatting per-Gaussian colors into an RGB image, it $\alpha$-blends these attributes into 2D material maps during rendering, and feeds them into the rendering equation to obtain per-pixel radiance with reflection:
\begin{align}
    L({\omega_o}) &= \int_\Omega L_i(\omega_i)f_r(\omega_i,\omega_o)\langle\omega_i \cdot \mathbf{n}\rangle\mathrm{d}\omega_i,
    \label{eq: burley}\\
    L_o &= L_d(\omega_o) + L_s(\omega_o).
\end{align}
Here $L_d$ and $L_s$ are the diffuse and specular radiances. $\omega_o$ and $\omega_i$ denote the outgoing and incident directions, respectively, determined by the viewpoint direction and surface normal $\mathbf{n}$, $L_i(\omega_i)$ is the incident radiance obtained by sampling a learnable environment map $E$, and $f_r$ is the BRDF that characterizes how much light is reflected by the object surface. Specifically, $f_r$ follows the Cook-Torrance model~\cite{cook1982reflectance} for $L_s$:
\begin{equation}
    f_r^s(\omega_i,\omega_o) = \frac{D\space G \space F}{4\langle\omega_o, \mathbf{n}\rangle\langle\omega_i,\mathbf{n}\rangle},
\end{equation}
and the Lambertian model $f_r^d = \boldsymbol{\lambda}/\pi$ for $L_d$. Refer to the \textbf{Supplementary Material} for details.

However, this formulation is unable to model the inter-reflection between two object surfaces. To this end, \emph{Ref-Gaussian} simply learns a spherical harmonics (SH) representation of indirect light for each Gaussian and uses it to replace DR-generated color for pixels not directly illuminated by the environment. The whole model is trained with the same set of losses as in vanilla 3DGS.

\section{Method}
\label{sec:method}

Our \method framework, as illustrated in Fig.~\ref{fig:pipeline}, features two main improvements upon \emph{Ref-Gaussian} that crucially enable polarization-based optimization and relighting for nonconvex objects. In Sec.~\ref{sec:pbrdf}, we detail our \polar process and its core pBRDF formulation. In Sec.~\ref{sec:envmap}, we introduce \subenvmap, an alternative strategy to model indirect light tailored for relighting tasks. 

\subsection{\polar: Polarimetric Deferred Rendering}
\label{sec:pbrdf}

The polarization state of light is represented by the \textbf{Stokes vector} $\mathbf{s} = \begin{bmatrix}s_0 \; s_1 \; s_2 \; s_3\end{bmatrix}^\top$, where $s_0$ denotes the total intensity, $s_1$ and $s_2$ represent the state of linear polarization, and $s_3$ characterizes circular polarization. 
Notably, changes to polarization during light-surface interaction can be modeled by a \textbf{Mueller matrix} $\mathbf{M}$ describing the surface's physical property, which maps the incident Stokes vector $\mathbf{s}_{\text{in}}$ to the outgoing one $\mathbf{s}_{\text{out}} = \mathbf{M}\mathbf{s}_{\text{in}}$.

According to the established pSVBRDF model~\cite{baek2018simultaneous, dave2022pandora}, specular reflection creates strong linear polarization from unpolarized incident light, while diffuse reflection results in much weaker polarization with angle of polarization (AoP) shifted by $90^\circ$. Therefore, ground truth polarization cues captured by either polarization camera or regular RGB camera overlaid with linear polarizers (LPs) encode rich information about the object's reflection property and thus they serve as an ideal modality to facilitate the learning of decomposed physical attributes. 

Following the formulation in pSVBRDF, the rendering equation Eq.~\ref{eq: burley} can be extended to a polarimetric form, \ie, \textbf{pBRDF}, which computes the polarization state of outgoing light after reflection. 
Assuming unpolarized incident light, specular reflection is modeled as:
\begin{equation}
\begin{gathered}
\label{eq:pbrdfspec}
    S_{\omega_o}^s =  \begin{bmatrix}
        1\\
        \beta_s(\theta_\mathbf{n})\cos{2\phi_\mathbf{n}}\\
        -\beta_s(\theta_\mathbf{n})\sin{2\phi_\mathbf{n}}
    \end{bmatrix} L_s(\omega_o), \\
     \text{where }
    \beta_s(\theta_\mathbf{n}) = \frac{R^\perp - R^\parallel}{R^\perp + R^\parallel},
    \end{gathered}
\end{equation}
and diffuse reflection is modeled as:
\begin{equation}
\begin{gathered}
    \label{eq:pbrdfdiff}
    S_{\omega_o}^d =  \begin{bmatrix}
        1\\
        \beta_d(\theta_\mathbf{n})\cos{2\phi_\mathbf{n}}\\
        -\beta_d(\theta_\mathbf{n})\sin{2\phi_\mathbf{n}}
    \end{bmatrix}L_d(\omega_o),\\
     \text{where} \,\, \beta_d(\theta_\mathbf{n}) = \frac{T^\perp - T^\parallel}{T^\perp + T^\parallel}.
\end{gathered}
\end{equation}
Herein, $L_d(\omega_o)$ and $L_s(\omega_o)$ denote outgoing radiance as formulated in Eq.~\ref{eq: burley}, and the Fresnel coefficients $T$ and $R$ denote the ratios of transmitted and reflected power to the incident light intensity, respectively, with ${\perp}$ and ${\parallel}$ indicating different orientations of linear polarization. All four Fresnel coefficients can be computed from the index of refraction (IoR) $\eta$ of the surface and the incident angle $\langle \mathbf{n}, \omega_i\rangle$. The zenith angle $\theta_\mathbf{n}$ is defined as $\theta_\mathbf{n} = \cos^{-1}(\mathbf{n} \cdot \omega_o)$, and the azimuth angle of the polarized light~\cite{baek2018simultaneous, hecht2012optics} $\phi_\mathbf{n}$ is computed as $\phi_\mathbf{n} = \cos^{-1}(\mathbf{n}_{\mathrm{cam}} \cdot \mathbf{y}_{\mathrm{cam}})$, where $\mathbf{y}_{\mathrm{cam}}$ and $\mathbf{n}_{\mathrm{cam}}$ are respectively the $y$-axis of the camera coordinate system and the projection of the surface normal $\mathbf{n}$ onto the image plane (perpendicular to the viewing direction). See Supplementary Material for a detailed formulation of pBRDF.

Similar to \emph{Ref-Gaussian}, our \polar process splats the following Gaussian attributes into material maps: albedo $\boldsymbol{\lambda}$, IoR $\eta$, surface normal $\mathbf{n}$, and roughness $r$. Specifically, $r$ is used to determine the mipmap level in which the learnable environment map $E$ is queried for specular incident radiance $L_i^s(\omega_i)$ computation. Then we follow Eqs.~\ref{eq:pbrdfspec}-\ref{eq:pbrdfdiff} to obtain $S_{\omega_o}^s$, $S_{\omega_o}^d$, and their sum $S_{\omega_o}$ as the overall rendered Stokes vector for each pixel, which is subsequently supervised by ground truth polarization information. Note that we refrain from using SH to represent color since albedo $\boldsymbol{\lambda}$ should be view-independent by definition.


\subsection{\subenvmap: Self-Occlusion-Aware Environment Map Building}
\label{sec:envmap}

We use an environment cube mipmap $E$ to represent incident light, whose pixel values stand for radiance from a specific direction at infinite distance. During \polar, $L_i^s$ and $L_i^d$ are obtained by querying $E$ with the incident direction $\omega_i$. However, when the target object is nonconvex, it may cast shadow on or even occlude itself, which results in inconsistency between queried and actual incident light, and creates unrealistic artifacts. This phenomenon is especially pronounced for the diffuse component $L_i^d$, since it integrates $E$ over a whole hemisphere and is more likely to be affected by self-occlusion.

On the one hand, \emph{Ref-Gaussian} circumvents this issue by learning a standalone SH representation of indirect light $\mathbf{l}_\mathrm{ind}$ for each Gaussian primitive. It then replaces DR-generated radiances with colors computed from $\mathbf{l}_\mathrm{ind}$ for any pixel not ``visible'' (\ie, the corresponding $\omega_i$ is blocked by the object itself) from the current viewpoint. Simple as this solution is, it fares poorly in relighting, since $\mathbf{l}_\mathrm{ind}$ simultaneously approximates global illumination and inter-reflection. Yet on the other hand, the classic approach to trace light through multiple reflection bounces till it hits the environment~\cite{chen2024gigs,gao2024r3dg} is also undesirable due to its high computation overhead during training.

\begin{figure}[t]
\centering
\includegraphics[width=\linewidth]{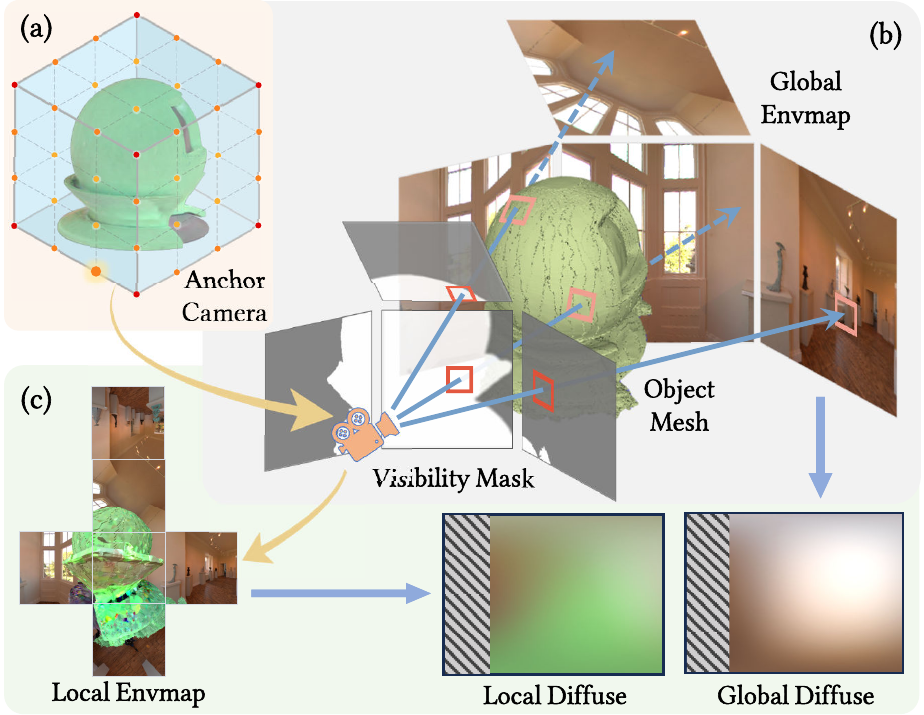} 
\vspace{-15pt}
\caption{Overview of \subenvmap. (a) We sample a set of anchor cameras on the object bounding box. (b) For each anchor camera, we ray trace in all directions to blend object color with the global environment map. (c) The resulting local environment map enables more accurate diffuse irradiance computation.} 
\vspace{-9pt}
\label{Fig:subenvmap} 
\end{figure}

Our \subenvmap finds balance between two extremes. As shown in Fig.~\ref{Fig:subenvmap}, we divide each face of the object's bounding box into 3$\times$3 grids and place an \textbf{anchor camera} at each grid point except for those on the bottom face. 
For each of the $N = 52$ cameras, we construct a \emph{local cubemap} $\tilde{E}_i$ with a resolution $D^2$ by performing $6\times D\times D$ one-step ray tracing, starting from the camera's location $\mathbf{c}_i$ and passing through each pixel on the cubemap.

When rendering for the pixel $(x, y)$, we perform one \polar step using each local cubemap as illumination to obtain a set of diffuse and specular Stokes, namely $\{\tilde{S}_d^{(1)}, \tilde{S}_d^{(2)}, \dots, \tilde{S}_d^{(N)}\}$ and $\{\tilde{S}_s^{(1)}, \tilde{S}_s^{(2)}, \dots, \tilde{S}_s^{(N)}\}$. Then, the \emph{localized diffuse Stokes} is computed as:
\begin{align}
\label{eq:subenvmap}
    \tilde{S}_d = \frac{\sum_{i=1}^N \|\mathbf{p} - \mathbf{c}_i\|_2\cdot\tilde{S}_d^{(i)}}{\sum_{i=1}^N \|\mathbf{p} - \mathbf{c}_i\|_2},
\end{align}
where $\mathbf{p}$ is the surface point corresponding to $(x, y)$. The \emph{localized specular Stokes} $\tilde{S}_s$ is calculated likewise, except that it only applies to ``invisible'' pixels as defined in prior work~\cite{yao2025refGS}. 
This design is necessary for optimizing $E$, since GridMaps are detached from the computation graph and $E$ cannot receive gradients from Eq.~\ref{eq:subenvmap} directly.

By blending object color with global illumination, each cubemap $\tilde{E}_i$ better captures the ``environment'' that surface points near the anchor $\mathbf{c}_i$ see, which may include the object itself. The distance-weighted formulation in Eq.~\ref{eq:subenvmap} further ensures that the object surface has smooth appearance and is free of abrupt color change. Since local cubemaps require no gradient and only need to be updated infrequently, the computation overhead of \subenvmap is mainly caused by the additional \polar steps, which is faster than performing multi-bounce ray tracing and easily parallelizable on GPUs.

\subsection{Training Strategies}
In all, the training loss of our model is formulated as:
\begin{equation}
\mathcal{L} = \mathcal{L}_{\mathrm{rgb}} + \lambda_1 \mathcal{L}_{\mathrm{pol}} + \lambda_2 \mathcal{L}_\mathrm{mask} + \lambda_3\mathcal{L}_\mathrm{depth} + \lambda_4 \mathcal{L}_{\mathrm{smooth}} \,.
\end{equation}

Following 3DGS, the RGB reconstruction loss is defined as $\mathcal{L}_{\mathrm{rgb}} = 0.8 \cdot\mathcal{L}_1(s_0,\hat{s}_0) + 0.2 \cdot\mathcal{L}_{\mathrm{DSSIM}}(s_0,\hat{s}_0)$. The polarization reconstruction loss is defined as $\mathcal{L}_{\mathrm{pol}} = \mathcal{L}_1(s_1,\hat{s}_1) + \mathcal{L}_1(s_2,\hat{s}_2)$. Here $\hat{s}_0$, $\hat{s}_1$ and $\hat{s}_2$ are the ground truth Stokes values.

Next, as we apply TSDF~\cite{yao2025refGS} to construct object mesh and use it for deferred rendering, floating Gaussians can be fatal to the reconstruction process. Therefore, we introduce a mask loss $\mathcal{L}_\mathrm{mask}$ to eliminate out-of-place floaters:
$ \mathcal{L}_\mathrm{mask} = \mathcal{L}_1 (M, \mathcal{O}) \, , $
where $\mathcal{O}$ denotes the rendered opacity map and $M$ is the per-frame segmentation mask given by pretrained segmentation models (for captured data) or the renderer (for synthetic data).

\begin{table*}[t]
\caption{Quantitative results of different methods on novel view synthesis (\textit{left}) and surface normal reconstruction (\textit{right}). Best results are highlighted as \colorbox{colorFst}{1st}, \colorbox{colorSnd}{2nd}, and \colorbox{colorTrd}{3rd}. This visualization applies to all following quantitative assessments.}
\vspace{-9pt}
\footnotesize
\label{tab:main_compare}
\centering
\renewcommand{\arraystretch}{1.3} 
\resizebox{\textwidth}{!}{%
\begin{tabular}{l||cc|cc|ccc|cc|cc|ccc|cc} 
\hline\hline
             & \multicolumn{9}{c|}{PSNR (dB) $\uparrow$}                    & \multicolumn{7}{c}{Cosine Distance (CD) $\downarrow$}                                                               \\ 
\hline
             & \multicolumn{2}{c|}{PANDORA}                              & \multicolumn{2}{c|}{RMVP} & \multicolumn{3}{c|}{SMVP} & \multicolumn{2}{c|}{Mitsuba3} & \multicolumn{2}{c|}{RMVP} & \multicolumn{3}{c|}{SMVP} & \multicolumn{2}{c}{Mitsuba3}  \\
             & owl & vase    & frog & dog                & squirrel & snail & david  & matpre. & teapot           & frog & dog                & squirrel & snail & david  & matpre. & teapot           \\ 
\hline
R3DG         & 24.49 & 25.35               & 26.56 & 17.77            & 19.71    & 26.97 & 22.93 & \cellcolor{colorSnd}25.41      & 28.30           & 0.0679 & 0.1376          & 0.0916   & 0.0473 & 0.1054 & 0.0795     & 0.0326           \\
GS-IR        & 23.76 & 24.40               & 31.07 & 33.33            & 18.59    & 27.92 & 20.47 & 16.43      & 21.77           & 0.0954 & 0.1523          & 0.0931   & 0.0909 & 0.1005 & 0.1130     & 0.0847           \\
GIR          & 23.81 & 23.37               & 28.26 &  \cellcolor{colorFst}41.10            & \cellcolor{colorTrd}20.88    & 23.67 & 22.78 & \cellcolor{colorTrd}24.41      & 27.10           & 0.0603 & 0.1674          & 0.0536   & 0.0491 & 0.1307 & \cellcolor{colorTrd}0.0616     & \cellcolor{colorTrd}0.0213           \\
3DGS-DR      & 24.20 & \cellcolor{colorTrd}26.94               &  \cellcolor{colorFst}34.68 & \cellcolor{colorSnd}39.59            & 20.63    & \cellcolor{colorSnd}28.46 & 24.72 & 24.33      & \cellcolor{colorTrd}29.07           & 0.0933 & 0.1420          & 0.0484   & 0.0462 & 0.1353 & 0.1069     & 0.0325 \\
Ref-Gaussian & 22.39 & 26.19               & \cellcolor{colorSnd}34.13 & \cellcolor{colorTrd}37.94            & 17.90    & 27.08 & 23.00 & 23.69      & \cellcolor{colorSnd}29.67           & \cellcolor{colorTrd}0.0501 & 0.1565          & 0.1122   & \cellcolor{colorTrd}0.0207 & 0.1067 & \cellcolor{colorSnd}0.0362     & \cellcolor{colorSnd}0.0093           \\
\hline
PolGS      & \cellcolor{colorTrd}24.99 & 24.85               & 28.25 & 28.15            & 20.92    & 26.23 & \cellcolor{colorTrd}27.20  &  -          &     -            & \cellcolor{colorSnd}0.0491 & \cellcolor{colorTrd}0.1085          & \cellcolor{colorTrd}0.0343   & 0.0297 & \cellcolor{colorTrd}0.0658 &   -         &    -              \\
PANDORA\footnotemark[1]      & \cellcolor{colorSnd!60}27.43 &  \cellcolor{colorFst!60}30.05               & 28.42 & 26.29            &  \cellcolor{colorFst!60}27.97    &  \cellcolor{colorFst!60}31.56 &  \cellcolor{colorFst!60}30.54 &      -      &         -        & 0.0533 & \cellcolor{colorSnd!60}0.0863          &  \cellcolor{colorFst!60}0.0154   & \cellcolor{colorSnd!60}0.0177 & 0.0663 &    -        &     -             \\
NeRSP \footnotemark[1]       &   -    &    -                 &  -     &   -               & \cellcolor{colorSnd!60}23.97    &  \cellcolor{colorFst!60}31.56 & \cellcolor{colorSnd!60}30.19 &     -       &   -              &   -     &   -              & \cellcolor{colorSnd!60}0.0243   &  \cellcolor{colorFst!60}0.0128 &  \cellcolor{colorFst!60}0.0545 &    -        &    -              \\
\hline
\textbf{Ours}     & \cellcolor{colorFst}28.14 & \cellcolor{colorSnd}27.39               & \cellcolor{colorTrd}32.92 & 37.82            & 20.01    & \cellcolor{colorTrd}28.18 & 24.62 &  \cellcolor{colorFst}26.56      &  \cellcolor{colorFst}29.69           &  \cellcolor{colorFst}0.0467 &  \cellcolor{colorFst}0.0910          & 0.0482   & 0.0261 & \cellcolor{colorSnd}0.0597 &  \cellcolor{colorFst}0.0334     &  \cellcolor{colorFst}0.0079    \\
\hline\hline
\end{tabular}
}
\vspace{-6pt}
\end{table*}

In addition, to make sure that 2DGS primitives align consistently with the object surface, we apply the depth-normal consistency loss $\mathcal{L}_\mathrm{depth}$ to enforce the $\alpha$-blended Gaussian normal $\mathbf{n}$ matches the depth normal $\tilde{\mathbf{n}}$, calculated from local depth gradients. That is:
$\mathcal{L}_\mathrm{depth} = 1 - \tilde{\mathbf{n}}^\top\mathbf{n} .$
Finally, $\mathcal{L}_{\mathrm{smooth}}=\|\nabla\mathbf{n}\|\exp{(-\|\nabla \hat{s}_0\|)}$ is an edge-aware normal smoothness loss, aiming to regularize normal variation.
\section{Experiments}
\label{sec:exp}

\subsection{Implementation Details}

\paragraph{Baselines.} We compare \method against both RGB-only and polarization-assisted models that are able to reconstruct reflective 3D objects. The former category includes GS-IR~\cite{Liang-gsir}, 3DGS-DR~\cite{ye2024gsdr}, Relightable3DGS (R3DG)~\cite{gao2024r3dg}, GIR\cite{shi2025gir}, and Ref-Gaussian~\cite{yao2025refGS}. 
The latter includes two NeRF-based methods, PANDORA~\cite{dave2022pandora} and NeRSP~\cite{han2024nersp}, as well as the concurrent work PolGS~\cite{han2025polgs}.

\begin{figure*}[h!]
\centering
\includegraphics[width=1\textwidth]{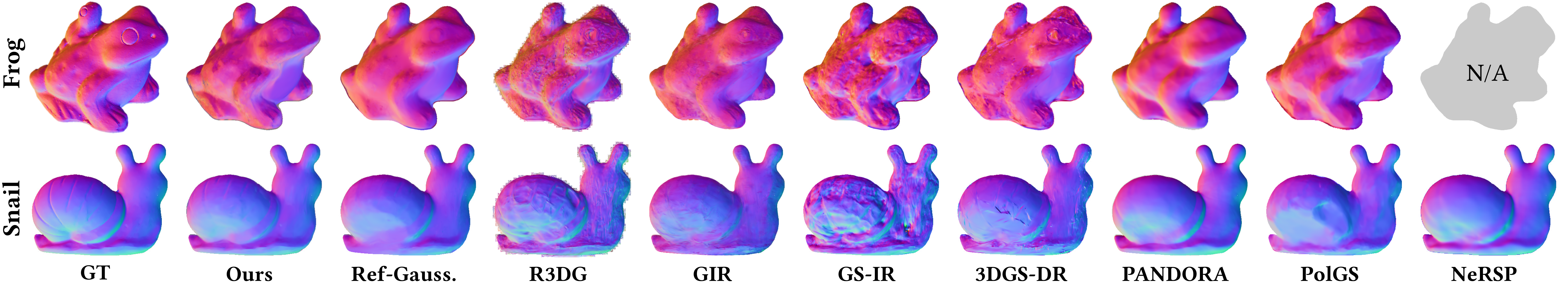}
\vspace{-14pt}
\caption{Visualization of reconstructed surface normal for synthetic (\textit{snail}) and real-world (\textit{frog}) objects.} 
\label{fig:normal}
\vspace{-0.3cm}
\end{figure*}

We train all GS-based models on an NVIDIA RTX 4090D GPU for 30k iterations. PANDORA and NeRSP models are trained on an NVIDIA V100 GPU for 150k and 50k iterations, respectively.

\vspace{-9pt}
\paragraph{Datasets.} We test all methods on 9 scenes from 4 datasets: two (\textit{owl} and \textit{vase}) from PANDORA~\cite{dave2022pandora}, two (\textit{frog} and \textit{dog}) from RMVP3D~\cite{han2024nersp}, three (\textit{squirrel}, \textit{snail} and \textit{david}) from SMVP3D~\cite{han2024nersp}, and two (\textit{matpre} and \textit{teapot}) rendered using Mitsuba3~\cite{jakob2022mitsuba3}. Among them, PANDORA and RMVP3D are captured datasets, while SMVP3D and Mitsuba3-rendered scenes are synthetic.

\footnotetext[1]{Training data for PANDORA and NeRSP models contain test viewpoints, and thereby their scores are only weakly comparable to others.}


\subsection{3D Reconstruction Results}

We evaluate 3D reconstruction quality from two aspects: novel view synthesis (NVS, measured by peak signal-to-noise ratio, or PSNR) and surface normal reconstruction (measured by Cosine distance, or CD). Table~\ref{tab:main_compare} displays a thorough quantitative comparison among different methods. More metrics such as SSIM and LPIPS for NVS and MAE for surface normal reconstruction are available in the Supplementary Material. 
We observe that our method exhibits overall better scores than RGB-based baselines and PolGS. While free of the conventional spherical harmonics (SH) representation of Gaussian color, \method is still able to match state-of-the-art methods in NVS, thanks to its accurate physically-grounded modeling of reflection. Moreover, polarization cues also assist with the learning of surface normal consistency, a trait that is hard to attain with GS variants. Our method is even comparable to PANDORA and NeRSP, trained with additional viewpoints for significantly more iterations, on several scenes.

A visual comparison of reconstructed surface normal is available in Fig.~\ref{fig:normal}. For results on other scenes and NVS visualizations, refer to the Supplementary Material.



\begin{figure}[t]
\centering
\includegraphics[width=\linewidth]{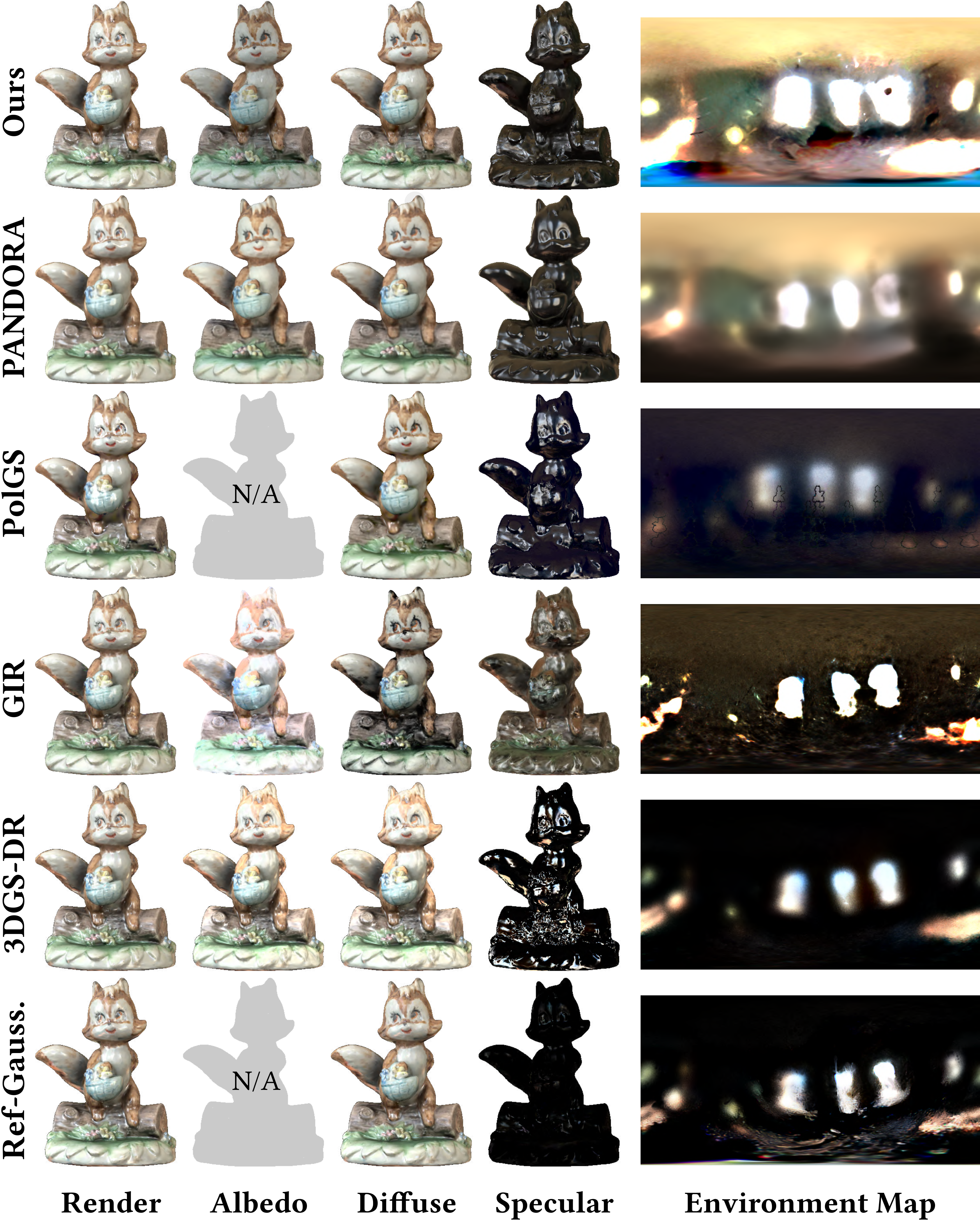}
\vspace{-15pt}
\caption{Comparison on reflection decomposition. Note that PolGS and Ref-Gaussian do not explicitly model object albedo.} 
\label{Fig:spec-diff_decomp}
\vspace{-6pt}
\end{figure}

\vspace{-9pt}
\paragraph{Reflection Decomposition.} By leveraging polarization information, \method achieves precise decomposition of reflection, including each pixel's albedo, diffuse radiance, specular radiance and the surrounding environment map, as visualized in Fig.~\ref{Fig:spec-diff_decomp}. Note that not all methods capable of reconstructing glossy objects learn explicit disentanglement of all relevant components. RGB-only methods, as we have predicted, are typically unable to resolve between the object's albedo and global illumination, while even the polarization-based PolGS is unable to correctly recover the brightness of the environment map.

\begin{figure*}[t]
\centering
\includegraphics[width=0.995\linewidth]{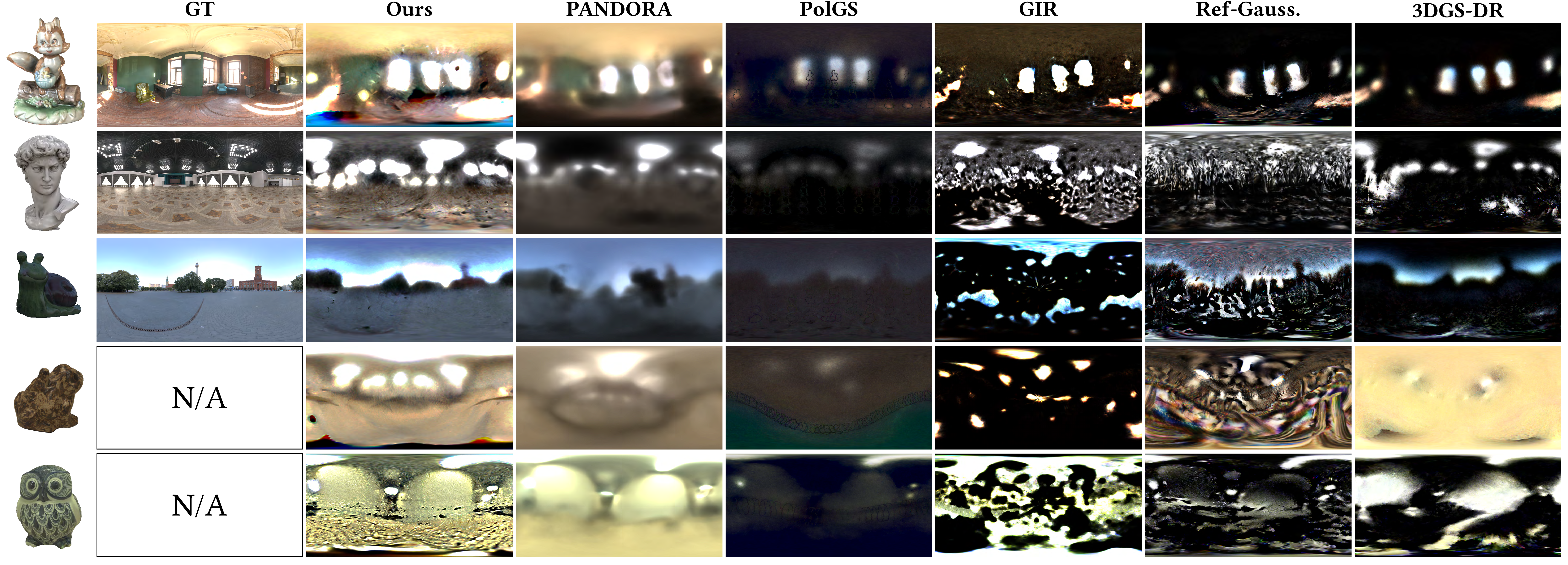}
\vspace{-9pt}
\caption{Qualitative comparison on estimated environment maps.} 
\label{Fig:envmap} 
\vspace{-6pt}
\end{figure*}

A more thorough comparison on environment map estimation is available in Fig.~\ref{Fig:envmap}. More results on \emph{owl} (from the PANDORA dataset) and \emph{frog} (from RMVP) are available in the Supplementary Material. 


\subsection{Relighting Results}

Relighting is a crucial task that determines whether the reconstructed 3D model can adapt to changes in scene content and subsequently be used in various downstream applications including VR/AR, gaming and interactive design. Unlike PANDORA that implicitly encodes environment maps or PolGS that does not decompose albedo, \method stands out as the first polarization-based reconstruction method that supports object relighting.

\begin{table}
\centering
\caption{Quantitative evaluation of environment map reconstruction and relighting. \textit{Left}: PSNR of optimized envmaps for different methods on the \emph{Mitsuba-rendered} dataset. \textit{Right}: Polarimetric evaluation of \textit{david} relighted with the \textit{sunset} envmap. Ablation results are also included. ``No PDR'' is realized by setting $\lambda_1 = 0$.}
\vspace{-6pt}
\scalebox{0.67}{%
\begin{tabular}{l||ccc|ccc} 
\hline\hline
\multirow{2}{*}{Method}    & \multicolumn{3}{c|}{Envmap (PSNR $\uparrow$)} & \multicolumn{3}{c}{Relighting}  \\ 
\cline{2-7}
                  & Teapot  & Matpre.   & David     & PSNR$\uparrow$  & SSIM$\uparrow$   & LPIPS$\downarrow$          \\ 
\hline
R3DG              & -     & -      & -          & 11.42 & 0.916 & 0.0754         \\
GS-IR             & 6.70 &	6.70	&6.72 & 16.90 & 0.957 & 0.0359         \\
GIR               & 10.30	&10.73	&10.01
    & \cellcolor{colorSnd}18.02 & \cellcolor{colorTrd}0.960 & \cellcolor{colorTrd}0.0327         \\
3DGS-DR           &6.71	&8.26&	6.71     & -     & -      & -              \\
Ref-Gaussian      & 7.51	&9.86	&8.28 & -     & -      & -              \\
\textbf{Ours} (no PDR \& GM)      & 9.97	&9.78&	10.69
     &15.56  &0.955  & 0.0369               \\
\textbf{Ours}  (no GM) & \textbf{12.17}	&15.82&	\textbf{13.00} & \cellcolor{colorTrd}17.81 & \cellcolor{colorSnd}0.967 & \cellcolor{colorSnd}0.0321         \\
\textbf{Ours}               & 11.50&	\textbf{17.46} &	12.39
     & \cellcolor{colorFst}19.18 & \cellcolor{colorFst}0.973 & \cellcolor{colorFst}0.0255         \\
\hline\hline
\end{tabular}
}
\vspace{-0.25cm}
\label{tab:relight_and_env}
\end{table}

\begin{figure}[t]
\centering
\includegraphics[width=\linewidth]{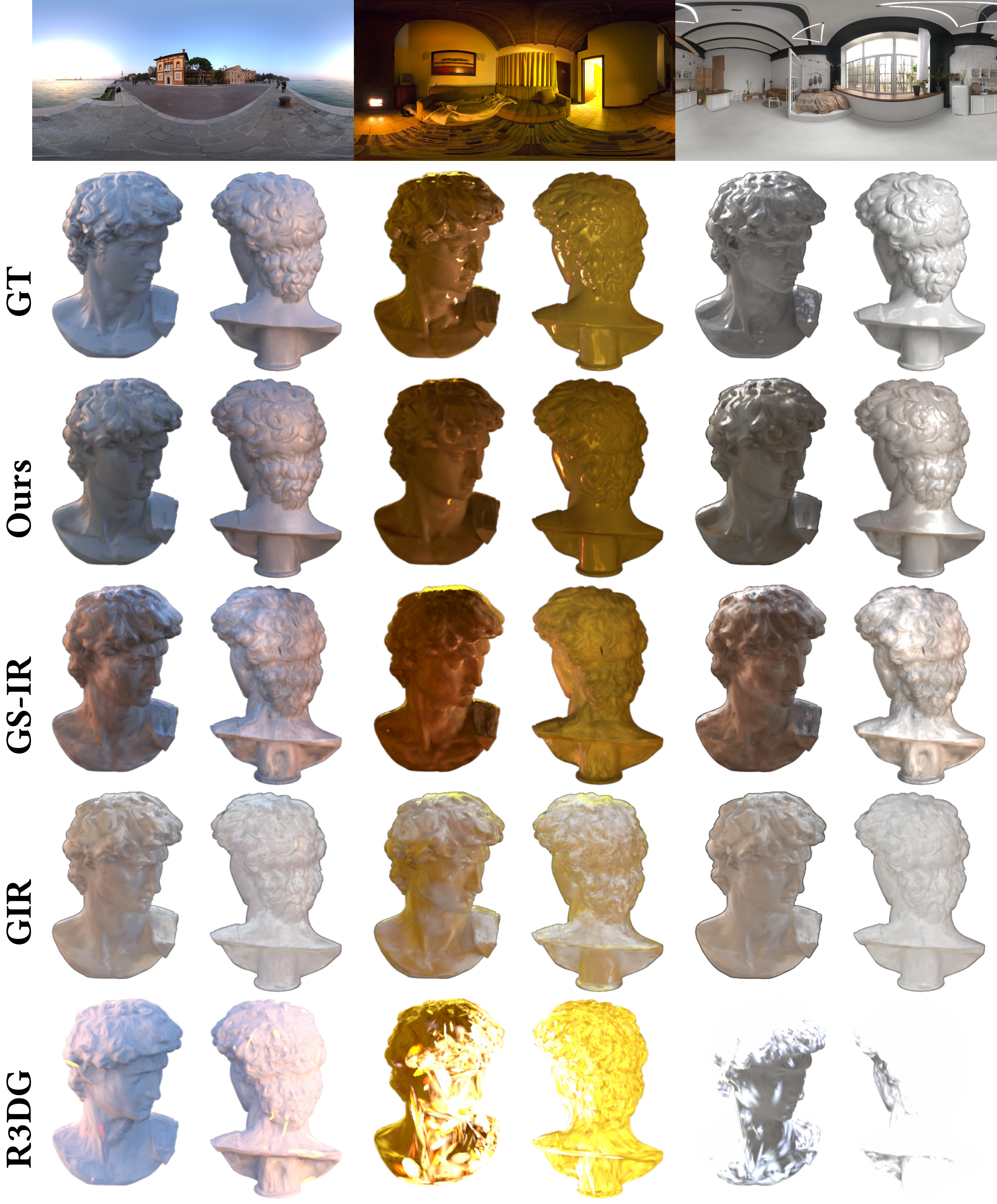}
\vspace{-16pt}
\caption{Relighting results from different methods. The environment maps are presented at the top. Note that R3DG suffers severe overexposure in the rightmost \textit{Brown Photostudio} environment.} 
\vspace{-14pt}
\label{Fig:relight} 
\end{figure}

We firstly compare the accuracy of environment map estimation in Table~\ref{tab:relight_and_env} \textit{left}. Three Mitsuba-rendered scenes where we have access to ground truth environment irradiance scales are used. Then, we test relighting by replacing the reconstructed environment maps and re-rendering the \emph{david} object. Quantitative and qualitative results are available in Table~\ref{tab:relight_and_env} \textit{right} and Fig.~\ref{Fig:relight}, respectively. Thanks to the high-fidelity albedo reconstructed by \polar and the occlusion-aware rendering enabled by \subenvmap, \method more faithfully preserves object color and simulates shades and highlights across different unseen environment maps.

\subsection{Reconstruction with Partial Polarization Cues}

To test our method's applicability in the absence of a delegated polarization camera, we additionally set up an acquisition system (see Fig.~\ref{fig:teaser}(a)) of two FLIR GS3-U3-23S6C-C RGB cameras, each overlaid with an off-the-shelf LBTEK FLP25-VIS-M linear polarizer (LP). Notably, this system only records \emph{partial} polarization information, yet as we can simulate LP by simply multiplying its corresponding Mueller matrix to the Stokes vectors calculated in Eqs.~\ref{eq:pbrdfspec}--\ref{eq:pbrdfdiff}, \method is readily trainable under such a setup. Examples on captured data and their reconstruction results are available in Fig.~\ref{Fig:device}. Details of the acquisition prototype and more results are available in the Supplementary Material.

\subsection{Ablation Study}

\begin{figure}[tbp]
\centering
\includegraphics[width=1.0\linewidth]{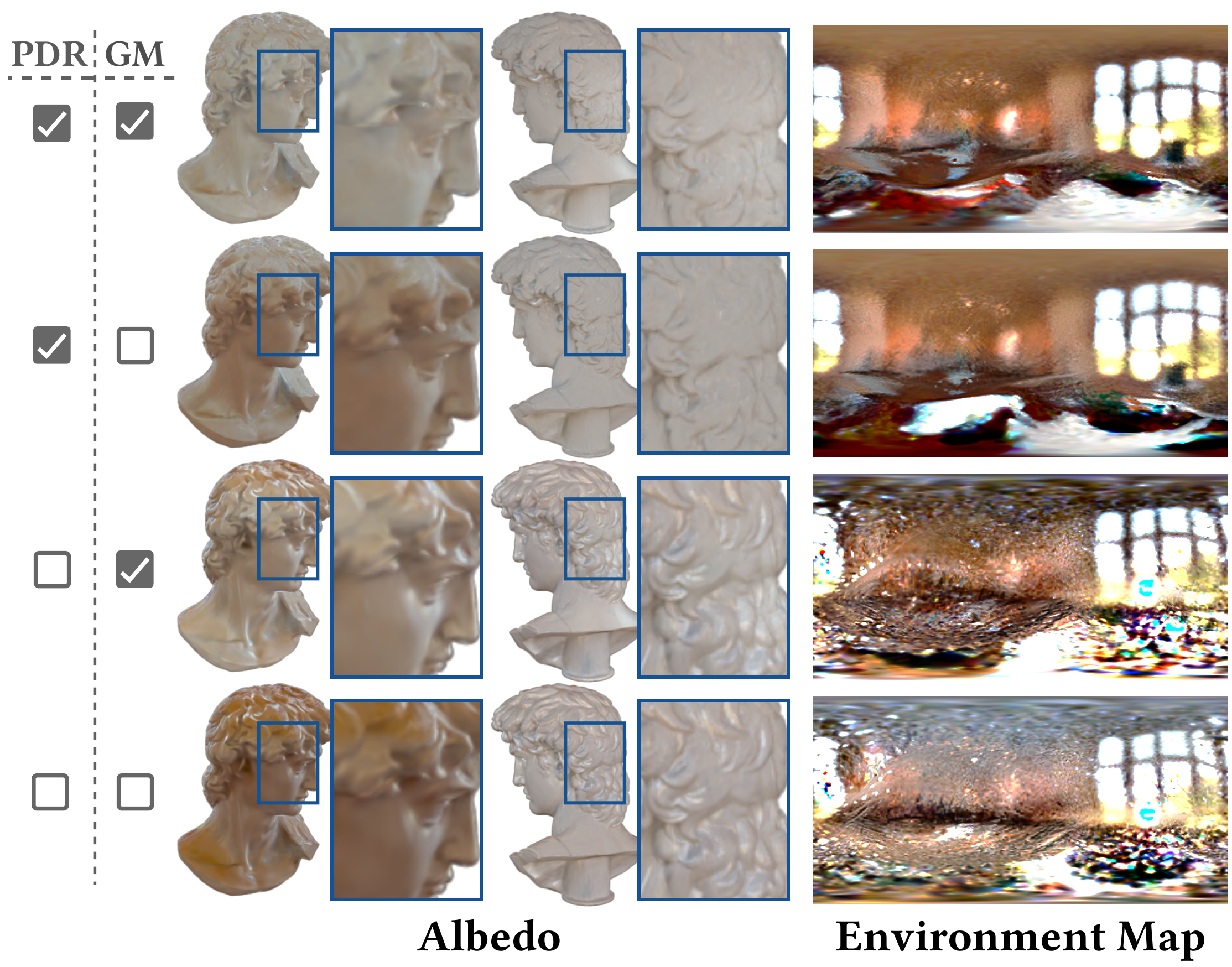}
\vspace{-18pt}
\caption{Ablation results with and without \polar (PDR) and/or \subenvmap (GM). We observe visible differences in color correctness, albedo quality, and environment map smoothness.} 
\label{Fig:ablation}
\vspace{-16pt}
\end{figure}

We conduct our ablation study on \textit{david} by controlling whether \polar and \subenvmap are used. As shown in Fig.~\ref{Fig:ablation}, polarization cues effectively exclude specular reflection from the reconstructed albedo, which consequently leads to significantly better environment map quality, while \subenvmap resolves shadows cast by the sculpture's non-convex geometry and helps to restore a consistent shade of whiteness across the surface of the object. Therefore, both components are essential for \method to achieve its established inverse rendering and relighting capabilities. Furthermore, quantitative results on environment map reconstruction and relighting are provided in Table~\ref{tab:relight_and_env}.

\section{Conclusion}
\label{sec:conclusion}

Unlike prior studies on 3DGS-based reconstruction relying solely on a sequence of multi-view RGB inputs, this work explores a robust polarization-based GS optimization pipeline for 3D reconstruction by incorporating physically-grounded \polar processing.
Furthermore, the presented \subenvmap mechanism successfully resolves indirect lighting and complex inter-reflection without learning and querying scene-specific parameters, substantially extending the relighting capability of \method to non-convex geometries and objects.
Experiments conducted on extensive synthetic and real-world data have demonstrated state-of-the-art performance in NVS and surface normal reconstruction of our \method, as well as its exceptional reflection decomposition and relighting capabilities, even when only partial polarization information is available.
We envision our proposed \method pipeline paves the way towards accurate and realistic 3D reconstruction of glossy objects, with broad applicability to VR/AR/MR and interactive designs.

\begin{figure}[t]
\centering
\includegraphics[width=0.975\linewidth]{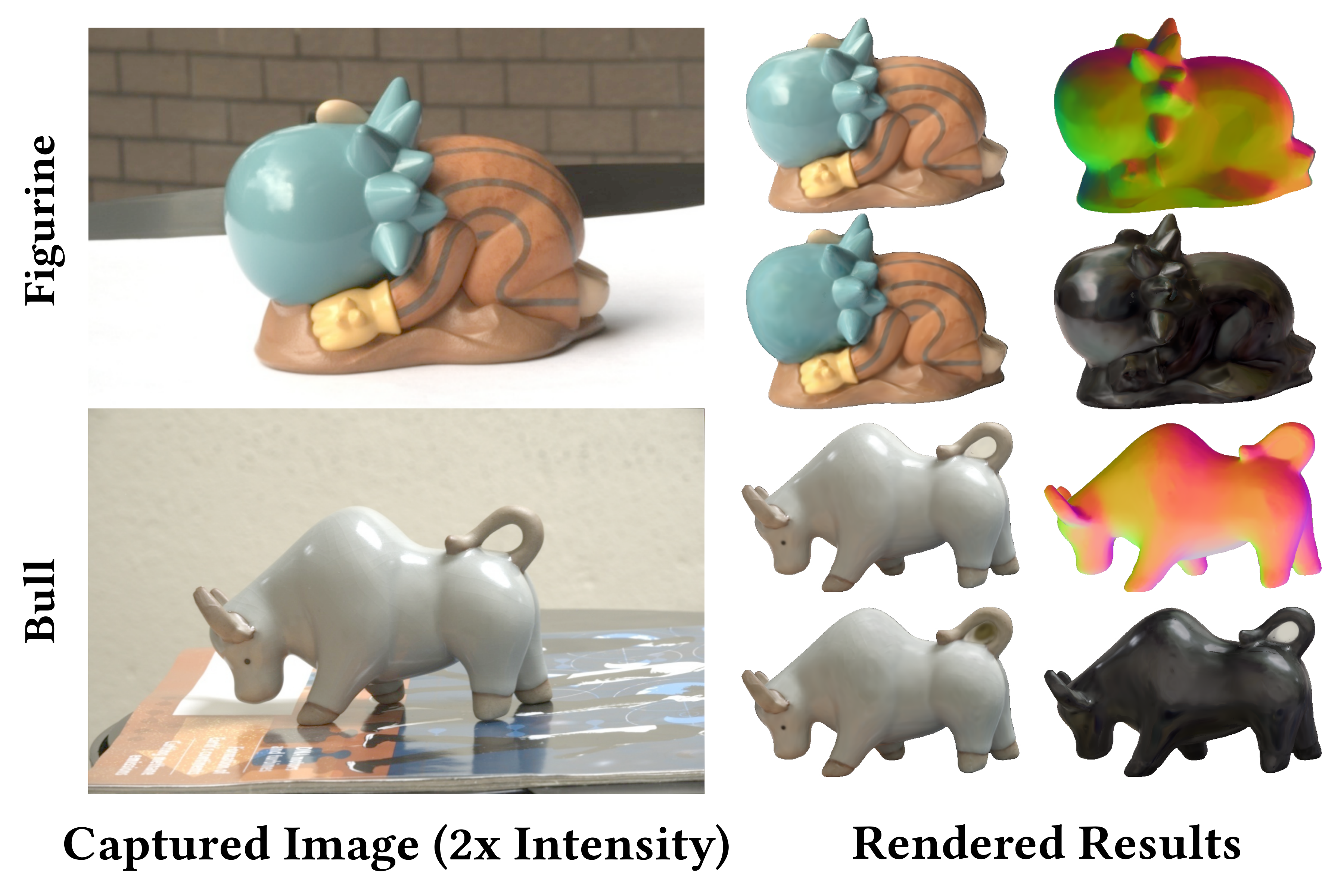}
\vspace{-9pt}
\caption{Examples of our real-world captured data and (\emph{from top-right, clockwise}) the reconstructed surface normal, specular reflection, diffuse reflection, and final rendered results.} 
\label{Fig:device} 
\vspace{-3pt}
\end{figure}

\vspace{-6pt}
\paragraph{Limitation and Future Works.}
The performance of \method may degenerate under certain circumstances.  Metal surfaces, whose pBRDFs involve complex-valued indices and phase terms, may lead to inaccurate reconstructions. 
Moreover, although \subenvmap effectively enhances existing envmap-based pipelines on global illumination modeling, it is still far from being the ultimate solution and may struggle with highly irregular object shapes or extremely specular surfaces that produce multiple inter-reflection bounces. We have presented failure cases and a comparison between different anchor placement strategies in the Supplementary Material. 
We also observe that reconstruction results on real-world data can be heavily affected by the disparity between actual 3D environments and our environment map formulation which assumes infinite depth of incident light. 

We have envisioned several key steps to tackle these limitations, including: (a) integrating metal surfaces into the PolarDR model; (b) modeling multi-bounce inter-reflection; (c) reducing device requirements of the PhyGaP framework; and (d) leveraging depth-aware environment representations, such as GaussProbe in TransparentGS~\cite{huang2025transparentgs}, to further boost the model's robustness and fidelity on real-world data.






\section*{Acknowledgement}

This work was partially supported by the National Key Research and Development Program of China (2024YFE0216600), the National Science Foundation of China (62322217, 62421003), the Innovation and Technology Fund of Hong Kong (MHP/313/24), and the Research Grants Council of Hong Kong (GRF 17208023).

{
    \small
    \bibliographystyle{ieeenat_fullname}
    \bibliography{main}
}
\newpage

\definecolor{colorTrd}{rgb}{0.95, 0.95, 0.75}
\definecolor{colorSnd}{rgb}{1, 0.85, 0.7}
\definecolor{colorFst}{rgb}{1, 0.7, 0.7}

\definecolor{matgreen}{rgb}{0.525, 0.894, 0.635}
\definecolor{matgray}{rgb}{0.463, 0.463, 0.463}

\definecolor{cvprblue}{rgb}{0.21,0.49,0.74}

\def\paperID{} 
\def\confName{CVPR}
\def\confYear{2026}

\def\method{PhyGaP\xspace} 




\maketitlesupplementary
\vspace{12pt}


\section{3D Gaussian Splatting (3DGS) Preliminaries}
\label{sec:preliminaries}

The established 3DGS \cite{kerbl3Dgaussians} represents a scene using 3D Gaussian Primitives $\mathcal{G} (\cdot)$, each characterized by a 3D mean position $\mathbf{\textbf{p}}$, a covariance matrix $\boldsymbol{\Sigma}$, an opacity $o$, and a set of spherical harmonics (SH) coefficients $\mathrm{SH}(\cdot)$ where view-dependent color $\mathbf{c}$ can be computed from viewpoint $\mathbf{r}$ as $\mathbf{c} = \mathrm{SH}(\mathbf{r})$.

To render an image, the 3D Gaussian is projected onto the camera space via:
\begin{align}
    p &= \mathbf{J}\mathbf{W}\mathbf{p},\\
    \Sigma_{\mathrm{2D}} &= \mathbf{J}\mathbf{W}\boldsymbol{\Sigma}\mathbf{W}^\top\mathbf{J}^\top,
\end{align}
where $p$ and $\Sigma_{\mathrm{2D}}$ are the 2D mean and the 2D covariance matrix respectively, $\mathbf{W}$ is the world-to-camera transformation matrix, and $\mathbf{J}$ is the Jacobian matrix of the perspective projection. The influence of each Gaussian at 2D coordinates $x$ is therefore defined as:
\begin{equation}
    \mathcal{G}_\mathrm{2D}(x) = \text{exp}\left[-\frac{1}{2}\left(x-p\right)^\top\Sigma_{\mathrm{2D}}^{-1}\left(x-p\right)\right].
\end{equation}
Next, volumetric $\alpha$-blending is employed for each pixel $x$ to blend $\alpha = o\cdot\mathcal{G}_\mathrm{2D}(x)$ in a front-to-back order to obtain the final color $C$, as follows:
\begin{equation}
    C = \sum_{i=1}^N c_i \cdot T_i \cdot \alpha_i, \text{where } T_i = \prod_{j=1}^{i-1}(1-\alpha_j).
    \label{eq:ab}
\end{equation}

Other physical attributes associated with each Gaussian primitive in works like \emph{Ref-Gaussian}~\cite{yao2025refGS}, such as roughness and surface normal, can be $\alpha$-blended into a feature map in a similar manner by replacing $c_i$ in Eq.~\ref{eq:ab} with the target attribute $a_i$.

\section{Detailed Formulation of PolarDR}
\subsection{Rendering Equation \& Split-Sum Approx.}
\label{supp:render_eq}
Based on Burley’s reflectance model \cite{2012PBRDisney, karis2013real} and Munkberg’s derivation \cite{munkberg2022extracting}, the reflected radiance $L_o(\omega_o)$ in direction $\omega_o$ is given by:
\begin{equation}
    \label{eq: burley}
    L_{\omega_o} = \int_\Omega L_i(\omega_i)f_r(\omega_i,\omega_o)\langle\omega_i, \mathbf{n}\rangle\mathrm{d}\omega_i .
\end{equation}

Notably, this equation integrates the product of incident light $L_i(\omega_i)$ from direction $\omega_i$ and the BRDF value $f_r(\omega_i,\omega_o)$. 
The integral is taken over the hemisphere $\Omega$ aligned with the surface normal $\mathbf{n}$.
The BRDF term $f_r(\omega_i,\omega_o)$ can be further decomposed into diffuse and specular components, $f_d$ and $f_s$, respectively. 



Following \emph{Ref-Gaussian}, we adopt the split-sum approximation \cite{karis2013real} to cope with the intractable integral of \textit{the specular component}:
\begin{align}
    L_s &= \int_{\Omega}f_s(\omega_i,\omega_o) L_s(\omega_i)\langle \omega_i ,\mathbf{n}\rangle\mathrm{d}\omega_i \notag\\
    &=\int_{\Omega} \frac{D\space G \space F}{4(\omega_o, \mathbf{n})(\omega_i,\mathbf{n})} L_s(\omega_i)\langle \omega_i ,\mathbf{n}\rangle\mathrm{d}\omega_i \\
    & \approx \underbrace{\int_{\Omega} \frac{D\space G \space F}{4(\omega_o, \mathbf{n})(\omega_i,\mathbf{n})}\langle \omega_i ,\mathbf{n}\rangle\mathrm{d}\omega_i}_{\text{Precomputed 2D Lookup }}\underbrace{\int_{\Omega} D L_s(\omega_i)\langle \omega_i ,\mathbf{n}\rangle\mathrm{d}\omega_i}_{\text{Environment Mip-Map}}\notag .
\end{align}

Note that \textbf{the first term} is independent of the incident light. Thus, we precompute and store a 2D lookup table indexed by the roughness $r$ and surface normal $\mathbf{n}$, which we can later query to realize $O(1)$ calculation of this term:
\begin{equation}
F_0 \cdot \tau_0 + \tau_1,    
\label{eq:2dlookup}
\end{equation}
where $\tau_{\{0, 1\}}$ are retrieved from the lookup table and $F_0$ refers to the Fresnel reflectance. For dieletric materials, $F_0$ is determined by the index of refraction (IoR) $\eta$:
\begin{equation}
    F_0 = \frac{(1-\eta)^2}{(1+\eta)^2}.
\end{equation}

\textbf{The second term} is represented with an environment cube mipmap $E(r, \omega)$ following Nvdiffrast~\cite{Laine2020diffrast} and Munkberg's implementation~\cite{munkberg2022extracting}. The base level possessing the minimum roughness $r$ ($0.08$ in our case) represents the pre-integrated lighting with highest resolution, while each subsequent mip-level with increasing $r$ is a filtered version of the previous one.

For \textit{the diffuse component}, the BRDF is given by $f_d = \boldsymbol{\lambda} / \pi$ following the Lambertian model, where $\boldsymbol{\lambda}$ is the albedo. However, since we only account for the light emits from \textbf{subsurface scattering (SSS)}, we need to deduct the portion of light reflected (and therefore does not participate in SSS). Following Schlick's model, the Fresnel term $F$ representing reflected energy is approximated by: 
\begin{equation}
    F = F_0 + (1-F_0)\cdot (1-\langle\omega'_i,\mathbf{n}\rangle)^5.
\end{equation}

 The term $\omega'_i$ here is the direction of incident light reflected from the surface, different from the integrated $\omega_i$. By excluding the diffuse pBRDF term $f_d$ from the integration, the integral can be reformulated as:
 
\begin{align}
    L_d(\mathbf{n}) &= (1 - F)f_d\int_{\Omega}L_d(\omega_i)\langle \omega_i ,\mathbf{n}\rangle\mathrm{d}\omega_i\notag\\
&\approx (1-F)f_d\sum_{ \scriptscriptstyle \langle\omega_i,\mathbf{n}\rangle > 0}L_{\mathrm{env}}(\omega_i) \langle\omega_i,\mathbf{n}\rangle ,
\label{eq:diffsup}
\end{align}
where $L_{\mathrm{env}}(\omega_i)$ is exactly $E(r_{\mathrm{min}}, \omega_i)$, that is, the environment map with the smallest roughness (\ie, highest resolution). Overall, the final pBRDF rendering equation can be expressed as:
\begin{align*}
    L_{\omega_o} = &(F_0 \tau_0+\tau_1)E(r,\omega_i) \\&+ \frac{\lambda}{\pi}(1-F)\sum_{ \scriptscriptstyle \langle\omega_i,\mathbf{n}\rangle > 0}E(r_\mathrm{min}, \omega_i) \langle\omega_i,\mathbf{n}\rangle.
\end{align*}

\subsection{Derivation of the pBRDF Model}

The Fresnel transmission matrix $\mathbf{F}$ is written as:
\begin{align}
    &\mathbf{F}^{F} = \notag\\ &\begin{bmatrix}\frac{F^\perp +F^\parallel}{2} & \frac{F^\perp -F^\parallel}{2} & 0 & 0 \\
    \frac{F^\perp -F^\parallel}{2} & \frac{F^\perp +F^\parallel}{2} & 0 & 0\\
    0 & 0 & \sqrt{F^\perp F^\parallel} \cos{\delta} & \sqrt{F^\perp F^\parallel} \sin{\delta}\\
     0 & 0 & -\sqrt{F^\perp F^\parallel} \sin{\delta} & \sqrt{F^\perp F^\parallel} \cos{\delta}
    \end{bmatrix},
\end{align}
where $F$ can be either Fresnel transmission coefficients $T$ or reflection coefficients $R$, and $\delta$ is the retardation phase shift between the perpendicular and parallel waves, either $\pi$ or $0$. $T$ and $R$ can be expressed as:
\footnotesize{\begin{align}
    &T^\perp = \left(\frac{2\eta_1 \cos{\theta_1}}{\eta_1 \cos{\theta_1} + \eta_2 \cos{\theta_2}}\right)^2,\\
    &T^\parallel = \left(\frac{2\eta_1 \cos{\theta_1}}{\eta_1 \cos{\theta_2} + \eta_2 \cos{\theta_1}}\right)^2,\\
    &R^\perp = \left(\frac{\eta_1 \cos{\theta_1} -\eta_2 \cos{\theta_2}}{\eta_1 \cos{\theta_1} + \eta_2 \cos{\theta_2}}\right)^2,\\
    &R_\parallel = \left(\frac{\eta_1 \cos{\theta_2} - \eta_2 \cos{\theta_1}}{\eta_1 \cos{\theta_2} + \eta_2 \cos{\theta_1}}\right)^2 ,
\end{align}}
\normalsize
where $\perp$ and $\parallel$ indicate perpendicular and parallel waves in transmitted ($T$) and reflected ($R$) light; $\eta_1$, $\eta_2$ are the indices of refraction (IoR) of the medium before and after the interface, which are set to 1.0 and the object IoR $\eta$, respectively; $\theta_1$ is the incident zenith angle, and $\cos{\theta_2} = \sqrt{1-(\frac{\eta_1}{\eta_2})^2\sin^2{\theta_1}}$, from Snell's law.

An interesting fact is that according to the energy conservation law we always have the following: 
\begin{equation}
    \frac{\eta_2 \cos{\theta_2}}{\eta_1 \cos{\theta_1}} T^{\perp/\parallel} + R^{\perp/\parallel} = 1.
\end{equation}

In order to transform Stokes vectors between the global coordinate system and the incident/exitant coordinate system, additional Mueller rotation matrices are applied before and after $\mathbf{F}$.
A complete derivation of this can be found in \cite{baek2018simultaneous}. The polarized shading models for diffuse and specular reflectance can be described as:
\begin{align}
&\mathbf{H}^d = \frac{\boldsymbol{\lambda}}{\pi} \langle\mathbf{n}, \mathbf{\omega}_i \rangle\cdot \notag\\
&
\begin{bmatrix}
 T_o^+ T_i^+ & T_o^+ T_i^- \beta_i & -T_o^+ T_i^- \alpha_i & 0\\
 T_o^- T_i^+ \beta_o & T_o^- T_i^- \beta_i \beta_o & -T_o^- T_i^- \alpha_i \beta_o & 0\\
 -T_o^- T_i^+ \alpha_o & -T_o^- T_i^- \alpha_o \beta_i & T_o^- T_i^- \alpha_i \alpha_o & 0\\
 0 & 0 & 0 & 0
\end{bmatrix},
\\
&\mathbf{H}^s = \frac{DG\langle \mathbf{n}, \mathbf{\omega}_i \rangle}{4\langle \mathbf{n},\mathbf{\omega}_o\rangle \langle \mathbf{n},\mathbf{\omega}_i \rangle}\cdot\notag\\
&\scalebox{0.85}{$
\left[
\begin{smallmatrix}
 R^+ & R^- \gamma_i & -R^- \gamma_i & 0\\
 R^- \gamma_o &
   R^+ \gamma_i \gamma_o + R^\times \chi_i \chi_o \cos\delta &
  -R^+ \chi_i \gamma_o + R^\times \gamma_i \chi_o \cos\delta &
   \chi_o R^\times \sin\delta\\
 -R^- \chi_o &
  -R^+ \gamma_i \chi_o + R^\times \chi_i \gamma_o \cos\delta &
   R^+ \chi_i \chi_o + R^\times \gamma_i \gamma_o \cos\delta &
   \gamma_o R^\times \sin\delta\\
 0 &
  -\chi_i R^\times \sin\delta &
  -\gamma_i R^\times \sin\delta &
   R^\times \cos\delta
\end{smallmatrix}
\right]
$}.
\end{align}
\normalsize
Here $T_i^\pm = \frac{T^\perp\pm T^\parallel}{2}$ represent the Fresnel transmission coefficients into the surface, and $T_o^\pm$ represent the Fresnel transmission coefficients out of the surface after SSS. Similarly, $R^\pm = \frac{R^\perp \pm R^\parallel}{2}$ where $R^+$ is the portion of energy reflected on the surface, and $R^\times$ is irrelevant to our model. 
$\alpha_{i,o}$ and $\beta_{i,o}$ denote $\sin(2\phi_{i,o})$ and $\cos(2\phi_{i,o})$, where $\phi_{i,o}$ are the azimuth angles of incident/outgoing light. Following the assumption of PANDORA~\cite{dave2022pandora}, the normal of microfacets $\mathbf{h}$ satisfies $\mathbf{h}=\mathbf{n}$, and $\chi_{i,o}$  and $\gamma_{i,o}$ are the same as $\alpha_{i,o}$ and $\beta_{i,o}$, respectively.

Specifically, for the diffuse $\mathbf{H}^d$, we suppose the scatter distance is small and no in-surface reflection take place, which means all energy participate in SSS will be emitted. Therefore, $T_o^+T_i^+$ can be equalized to the $1 - F$ term in Eq.~\ref{eq:diffsup} since they both represent the ratio of light participating in SSS. Similarly, $R^+$ can be equalized to the ratio of reflected light, \ie $F$ in Section~\ref{supp:render_eq}.

Suppose that incident light is unpolarized with Stokes vector $S_{in} = \begin{bmatrix}L_{\omega_i}&0&0&0\end{bmatrix}^\top$, the diffuse component of the outgoing Stokes vector can then be computed as:
\begin{align}
    S_{out}^d(\mathbf{\omega}_i) &=\mathbf{H}^d \cdot \begin{bmatrix}L_{\omega_i}&0&0&0\end{bmatrix}\notag\\ &= \frac{\boldsymbol{\lambda}}{\pi} \langle\mathbf{n}, \mathbf{\omega}_i \rangle\begin{bmatrix}
         T_o^+ T_i^+\\
         T_o^- T_i^+ \beta_o\\
         -T_o^- T_i^+ \alpha_o\\
         0
    \end{bmatrix} \cdot L_{\omega_i}\\
    &= (1-F)\cdot\frac{\boldsymbol{\lambda}}{\pi} \langle\mathbf{n}, \mathbf{\omega}_i \rangle \cdot\begin{bmatrix}
         1\\
         (T_o^- /T_o^+)\beta_o\\
         (-T_o^- / T_o^+) \alpha_o\\
         0
    \end{bmatrix} \cdot L_{\omega_i}.\notag
\end{align}


Similarly, the specular component is:
\begin{align}
    S_{out}^s (\mathbf{\omega}_i) &= \mathbf{H}^s \cdot \begin{bmatrix}L_{\omega_i}&0&0&0\end{bmatrix}\notag\\
    &= \frac{DGF}{4\langle \mathbf{n},\mathbf{\omega}_o\rangle \langle \mathbf{n},\mathbf{\omega}_i \rangle} \langle \mathbf{n}, \mathbf{\omega}_i \rangle \begin{bmatrix}
         1\\
        (R^-/R^+) \beta_o\\
        (R^-/R^+) \alpha_o\\
         0
    \end{bmatrix} \cdot L_{\omega_i}.
\end{align}

Please refer to the following papers for more details: Mitsuba3~\cite{jakob2022mitsuba3}, PANDORA~\cite{dave2022pandora}, and pSVBRDF~\cite{baek2018simultaneous}.

\begin{figure*}[t]
\centering
\includegraphics[width=\textwidth]{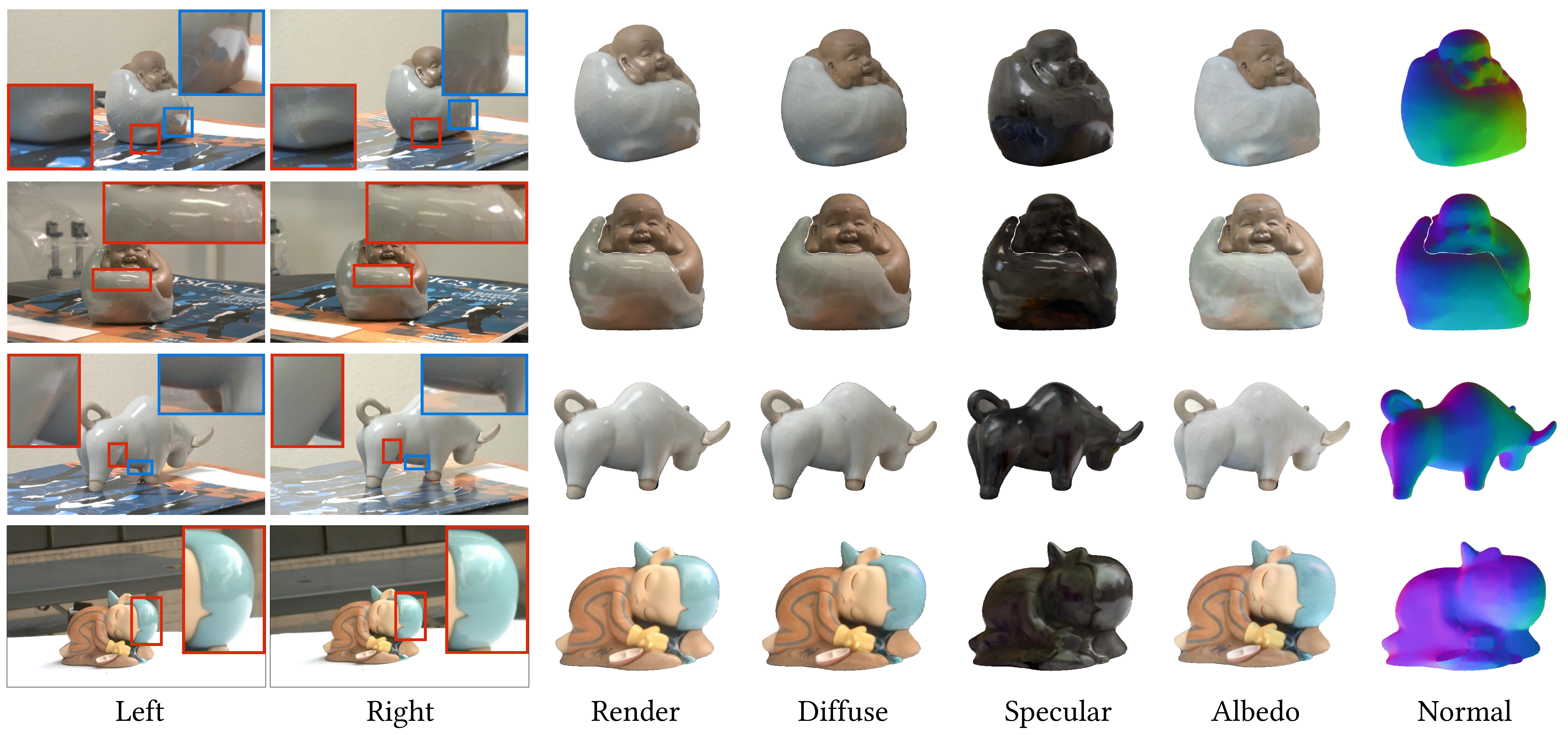}
\caption{Real-word acquisition samples and reconstruction results. Note that reflection patterns captured by two cameras differ, due to the angular disparity between two LPs. From top to bottom: \textit{buddha-room}, \textit{buddha-corridor}, \textit{bull-corridor}, and \textit{figurine-garden}.} 
\label{fig:Captured Data} 
\end{figure*}

\begin{table*}[t]
\caption{SSIM$\uparrow$ (\textit{left}) and LPIPS$\downarrow$ (\textit{right}) of different methods on novel view synthesis. Best and second best results are indicated by \textbf{bold} and \underline{underline} fonts respectively.}
\footnotesize
\label{tab:supp_ssim_lpips}
\centering
\renewcommand{\arraystretch}{1.3} 
\resizebox{\textwidth}{!}{%
\begin{tabular}{l||cc|cc|ccc|cc} 
\hline\hline
         \multirow{ 2}{*}{Methods}    & \multicolumn{2}{c|}{PANDORA}                              & \multicolumn{2}{c|}{RMVP} & \multicolumn{3}{c|}{SMVP} & \multicolumn{2}{c}{Mitsuba3}\\
             \cline{2-10}
             & owl & vase    & frog & dog                & squirrel & snail & david  & matpre. & teapot\\ 
\hline
R3DG         & 0.935 / 0.059 & 0.962 / 0.052 & 0.983 / 0.032 & 0.955 / 0.029 & 0.915 / 0.071 & 0.932 / 0.098 & 0.939 / 0.062 & 0.968 / 0.047 & 0.945 / 0.147 \\
GS-IR        & 0.915 / 0.076 & 0.929 / 0.091 & 0.983 / 0.028 & 0.996 / 0.011 & 0.898 / 0.088 & 0.939 / 0.096 & 0.921 / 0.085 & 0.903 / 0.085 & 0.882 / 0.179\\
GIR          & 0.931 / 0.058 & 0.963 / 0.053 & 0.987 / \underline{0.023} & \textbf{0.999} / \textbf{0.003} & 0.936 / 0.058 & 0.940 / 0.090 & 0.949 / 0.051 & 0.966 / 0.051 & 0.941 / 0.149 \\
3DGS-DR      & 0.938 / 0.050 & 0.971 / 0.040 & \textbf{0.991} / \textbf{0.018} & \textbf{0.999} / \underline{0.006} & 0.933 / 0.057 & 0.947 / 0.075 & 0.952 / 0.043 & 0.968 / 0.046 & 0.947 / \underline{0.135}\\
Ref-Gaussian & 0.924 / 0.064 & 0.972 / \underline{0.039} & \underline{0.988} / 0.028 & \textbf{0.999} / 0.008 & 0.897 / 0.084 & 0.941 / 0.079 & 0.950 / 0.049 & \underline{0.971} / \underline{0.043} & \underline{0.949} / \underline{0.135} \\
\hline
PolGS      & \underline{0.941} / 0.064 & 0.960 / 0.064 & 0.965 / 0.043 & 0.991 / 0.022 & 0.926 / 0.071 & 0.921 / 0.126 & 0.966 / \underline{0.038} &  -          &     -    \\
PANDORA\footnotemark[1]      & \color{gray}\textbf{0.960} / \textbf{0.042} & \color{gray}\textbf{0.984} / \textbf{0.038} & \color{gray}0.983 / 0.026 & \color{gray}\textbf{0.997} / 0.008 &  \color{gray}\textbf{0.966} / \textbf{0.042} &  \color{gray}\textbf{0.976} / \underline{0.068} & \color{gray}\underline{0.968} / 0.051 &      -      &         -        \\
NeRSP \footnotemark[1]       &   -    &    -                 &  -     &   -               & \color{gray}\underline{0.958} / \underline{0.049} & \color{gray}\underline{0.975} / \textbf{0.059} & \color{gray}\textbf{0.979} / \textbf{0.030} &     -       &   -              \\
\hline
\textbf{Ours}     & \textbf{0.960} / \underline{0.043} & \underline{0.975} / 0.048 & \underline{0.988} / 0.024 & \textbf{0.999} / \underline{0.006} & 0.920 / 0.070 & 0.943 / 0.094 & 0.955 / 0.040 & \textbf{0.977} / \textbf{0.042} &  \textbf{0.957} / \textbf{0.131} \\
\hline\hline
\end{tabular}
}
\end{table*}

\begin{figure*}[!tp]
\centering
\includegraphics[width=1\textwidth]{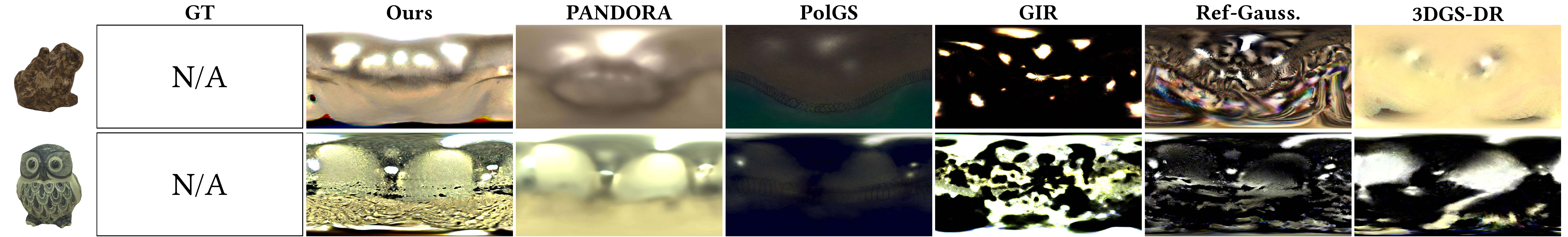}
\caption{More qualitative comparison on estimated environment maps. Ground truths for these two scenes (\textit{frog} and \textit{owl}) are not available.} 
\label{fig:supp_envgrid}
\vspace{-0.3cm}
\end{figure*}

\subsection{Training with Partially Polarization Cues}

In cases where a polarization camera is unavailable, we may alternatively use regular RGB cameras overlaid with linear polarizers (LP) to acquire polarization information. The Muller Matrix of LP with orientation $\theta$ is written as: 
\begin{equation}
    \mathbf{M}_{\mathrm{LP}}(\theta) = \frac{1}{2} \begin{bmatrix}
        1&\cos2\theta&\sin2\theta&0\\
        \cos2\theta&\cos^22\theta&\cos2\theta\sin2\theta&0\\
        \sin2\theta&\cos2\theta\sin2\theta&\sin^22\theta&0\\
        0&0&0&0
    \end{bmatrix}.
\end{equation}

Therefore, we may formulate our Stokes loss between rendered Stokes $S_{out}$ and captured light intensity $I$ as:
\begin{equation}
    \mathcal{L} = \|[\mathbf{M}_{\mathrm{LP}}(\theta)\cdot S_{out}]_0 - I\|_1.
\end{equation}






Note that $\theta$ is not necessarily known, and can be optimized alongside other parameters. A more detailed analysis can be found in Wu et al.'s work~\cite{wu2025glossy}.

\section{Training Details}
\subsection{Training Parameters}

For most of our scenes with full polarization information, we set $\lambda_1 = 10$ and $\lambda_3 = 0.2$, except for \textit{frog} and \textit{dog} in the RMVP dataset, that we set $\lambda_1 = 1$ and $\lambda_3 = 0.3$. For all scenes, we set $\lambda_2 = 0.4$ and $\lambda_4 = 0.1$. For the index of refraction (IoR) $\eta$, we apply the activation function $\sigma(\mu) = \mathrm{Sigmoid}(\mu) + 0.3$ to map it to $(1.3, 2.3)$, as $1.3$ is the IoR of water and dielectric materials typically have IoR between $1.5$ and $2$. 

\subsection{Input Data}
We convert all input images to linear color space to keep consistent with the rendering model. All results are converted to the sRGB color space for visualization. 
For GS-based methods, we hold out $1/8$ of all viewpoints as the test set for each scene.

\subsection{Environment Map Details}
\label{supp:envmap}
We assume that environment maps encode light intensity at each pixel and therefore dwells in the linear color space. Consequently, we use the following activation function:
\begin{equation}
f(x) = 
  \begin{cases}
    \mathrm{Sigmoid}(x) , & x \le 0\\
    x + 0.5 , & x > 0
  \end{cases}.
\end{equation}

All environment maps are converted to the sRGB color space for visualization.

\subsection{GridMap Details}

We update GridMap every 300 iterations to reflect changes in global illumination without causing too much computation overhead. The measured time overheads of GridMap are 24.1\% and 26.8\% for training and inference, respectively. We set $D = 64$ for cubemap construction. 

In each one-step ray tracing that touches the object mesh, we retrieve the intersection point $\mathbf{p}$ and the surface normal $\mathbf{n}'$ of the mesh at $\mathbf{p}$. We perform the kNN algorithm to obtain an approximated albedo $\boldsymbol{\lambda}'$ at $p$, and use $\mathbf{n}'$ to query the global environment for global diffuse illumination $L_d(\mathbf{n}')$. Then, we calculate the local outgoing intensity of the object at $p$ as $\mathbf{c} = \frac{\boldsymbol{\lambda'}}{\pi}L_d(\mathbf{n}')$ (essentially Eq.~\ref{eq:diffsup} without the Fresnel term). For rays that miss the object mesh, we directly copy the pixel values of the base environment map to construct the local cubemap.

We use a slightly larger bounding box, \ie, $1.1\times$, for GridMap construction to avoid placing anchor cameras inside the object.

\subsection{Data Acquisition Prototype Details}
Our acquisition system features two RGB cameras, each overlaid with an linear polarizer (LP). The baseline between two cameras is set around 10~cm, and the LPs are rotated to approximately horizontal and vertical positions. 
During training, the rotation angles of the LPs are optimized to $170^\circ$$\sim$$10^\circ$ and $80^\circ$$\sim$$100^\circ$, respectively. 
We capture four scenes using this setup: \textit{buddha-room}, \textit{buddha-corridor}, \textit{bull-corridor}, and \textit{figurine-garden}, featuring objects of different materials (porcelain vs. plastic) and different environments (both indoor and outdoor). 
Each scene contains 50--80 viewpoints (\ie, 100--160 images in total) that are uniformly sampled along a circular trajectory with radius $\sim$60~cm around the object. 
Noteworthy, careful readers may notice in the main text visualization that we set up an extra RGB camera (w/o LP) on our prototype rig, which is only to acquire data for baseline comparison matter.

The original image resolutions are $1,920\times1,200$, which we downsample to $1,152\times720$ during training. We apply the well-established COLMAP algorithm~\cite{schonberger2016structure} to estimate camera poses, and use the langSAM and SAM~\cite{kirillov2023segment} models to obtain object masks. Figure~\ref{fig:Captured Data} visualizes reconstructed results of each scene using our PhyGaP framework.


\section{Additional Results}

Table~\ref{tab:supp_ssim_lpips} compares the SSIM$\uparrow$ and LPIPS$\downarrow$ scores of different methods on NVS, and Table~\ref{tab:supp_mae} compares the MAE$\downarrow$ scores on surface normal reconstruction.

\begin{figure*}[t]
\centering
\includegraphics[width=\textwidth]{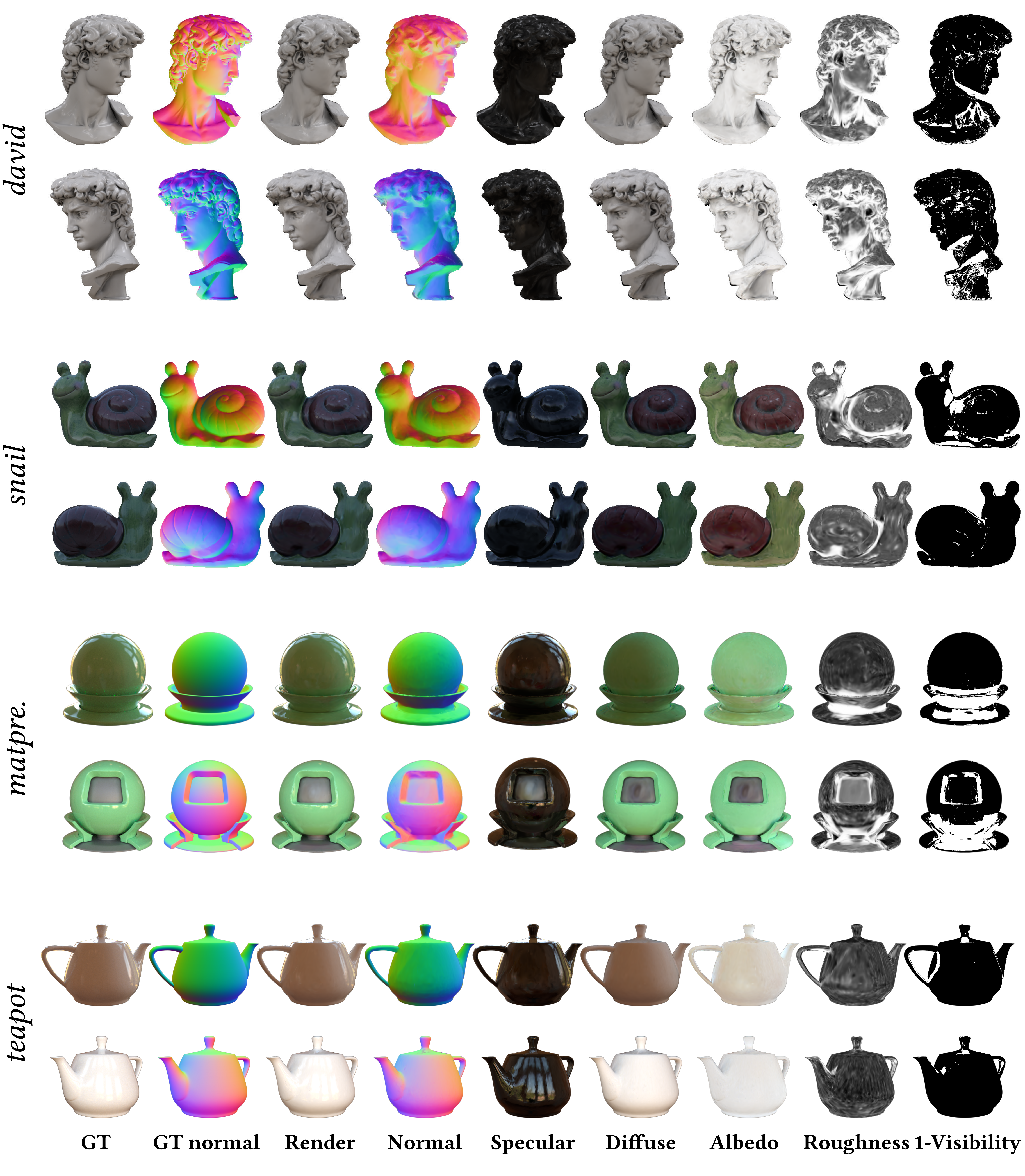}
\vspace{-9pt}
\caption{Additional reconstruction results with our model. From top to bottom: \textit{david}, \textit{snail}, \textit{matpre.}, and \textit{teapot}.} 
\label{Fig:more results} 
\vspace{16pt}
\end{figure*}

Figure~\ref{Fig:more results} visualizes additional reconstruction results of our PhyGaP method for different scenes. We observe that two RGB-LP cameras capture different reflection patterns, which enables our model to effectively decompose diffuse, specular and albedo, as well as to reconstruct smooth and realistic surface normals.

Additional comparisons on environment map estimation are available in Figure~\ref{fig:supp_envgrid}. 

\newpage

\subsection{Evaluating Geometry Reconstruction}

In addition to Cosine Distance and MAE that only validates surface normal consistency, we also measure Chamfer Distance (CD) to directly evaluate geometry reconstruction quality. 
For fair comparison, we firstly unproject the rasterized depth $d$ of each pixel into the 3D space as $\mathbf{p} = d\cdot\mathbf{v} + \mathbf{p}_{\mathrm{cam}}$, where $\mathbf{v}$ is the camera viewing ray and $\mathbf{p}_\mathrm{cam}$ is the camera center, and then use Poisson surface reconstruction with depth $8$ and pruning threshold $5\times10^{-3}$ to generate meshes. 

We compare our method against three GS-based baselines that produce relatively high-quality meshes with this process, namely: PolGS~\cite{han2025polgs}, GS-IR~\cite{Liang-gsir}, and GIR~\cite{shi2025gir}. Unidirectional (from predicted mesh to ground truth mesh, within the bounding box of the ground truth mesh) L1 distances are calculated.

\begin{figure}[t]
    \centering
    \includegraphics[width=0.95\linewidth]{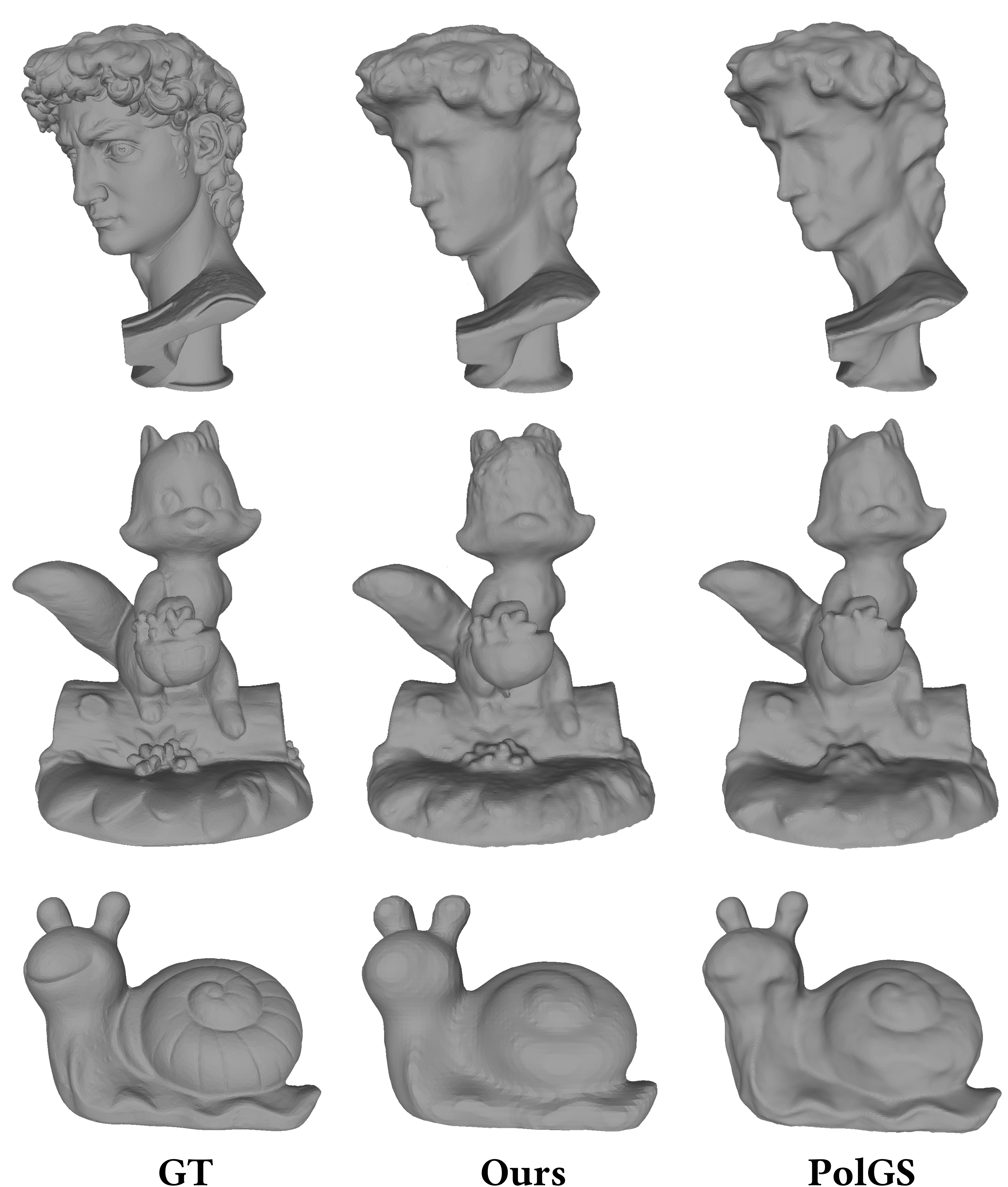}
    \caption{Meshes reconstructed by PhyGaP and PolGS. Results for GIR and GS-IR are not presented since their meshes exhibit significantly worse quality due to floating Gaussian points.}
    \label{fig:mesh_viz}
\end{figure}

As shown in Table~\ref{supp:mesh_metric} and Figure~\ref{fig:mesh_viz}, we achieve comparable results as the previous state-of-the-art, \ie, PolGS, and beats GIR and GS-IR by a large margin.

\begin{table}[t]
\caption{MAE$\downarrow$ of different methods on surface normal reconstruction. Best and second best results are indicated by \textbf{bold} and \underline{underline} fonts respectively.}
\vspace{-6pt}
\footnotesize
\label{tab:supp_mae}
\centering
\renewcommand{\arraystretch}{1.3} 
\resizebox{\linewidth}{!}{%
\begin{tabular}{l||cc|ccc|cc} 
\hline\hline
        \multirow{ 2}{*}{Methods}     &  \multicolumn{2}{c|}{RMVP} & \multicolumn{3}{c|}{SMVP} & \multicolumn{2}{c}{Mitsuba3}\\
             \cline{2-8}
             & frog & dog                & squirrel & snail & david  & matpre. & teapot\\ 
\hline
R3DG         & 17.46 & 24.86 & 19.90 & 13.22 & 21.12 & 18.22 & 11.62 \\
GS-IR        & 21.35 & 27.31 & 21.13 & 20.86 & 21.62 & 22.45 & 19.26\\
GIR          & 15.98 & 29.74 & 13.91 & 14.04 & 24.18 & 13.61 & 7.90 \\
3DGS-DR      & 19.89 & 24.96 & 12.01 & 13.03 & 24.06 & 20.64 & 9.21\\
Ref-Gaussian & \underline{14.01} & 27.43 & 17.12 & \underline{8.35} & 21.23 & \underline{9.01} & \underline{4.97} \\
\hline
PolGS      & 14.27 & 21.85 & 9.91 & 10.35 & 16.02 &  -          &     -    \\
PANDORA\footnotemark[1]      & \color{gray}14.28 & \color{gray}\textbf{18.04} & \color{gray}\textbf{5.91} & \color{gray}18.45 & \color{gray}15.51 &      -      &         -        \\
NeRSP \footnotemark[1]       &  -     &   -               & \color{gray}\underline{8.02} & \color{gray}\textbf{5.52} & \color{gray}\underline{13.99} &     -       &   -              \\
\hline
\textbf{Ours}     & \textbf{13.58} & \underline{19.11} & 13.51 & 9.49 & \textbf{13.72} & \textbf{8.32} & \textbf{4.03} \\
\hline\hline
\end{tabular}
}
\vspace{6pt}
\end{table}

\begin{figure}[h]
    \centering
    \includegraphics[width=0.95\linewidth]{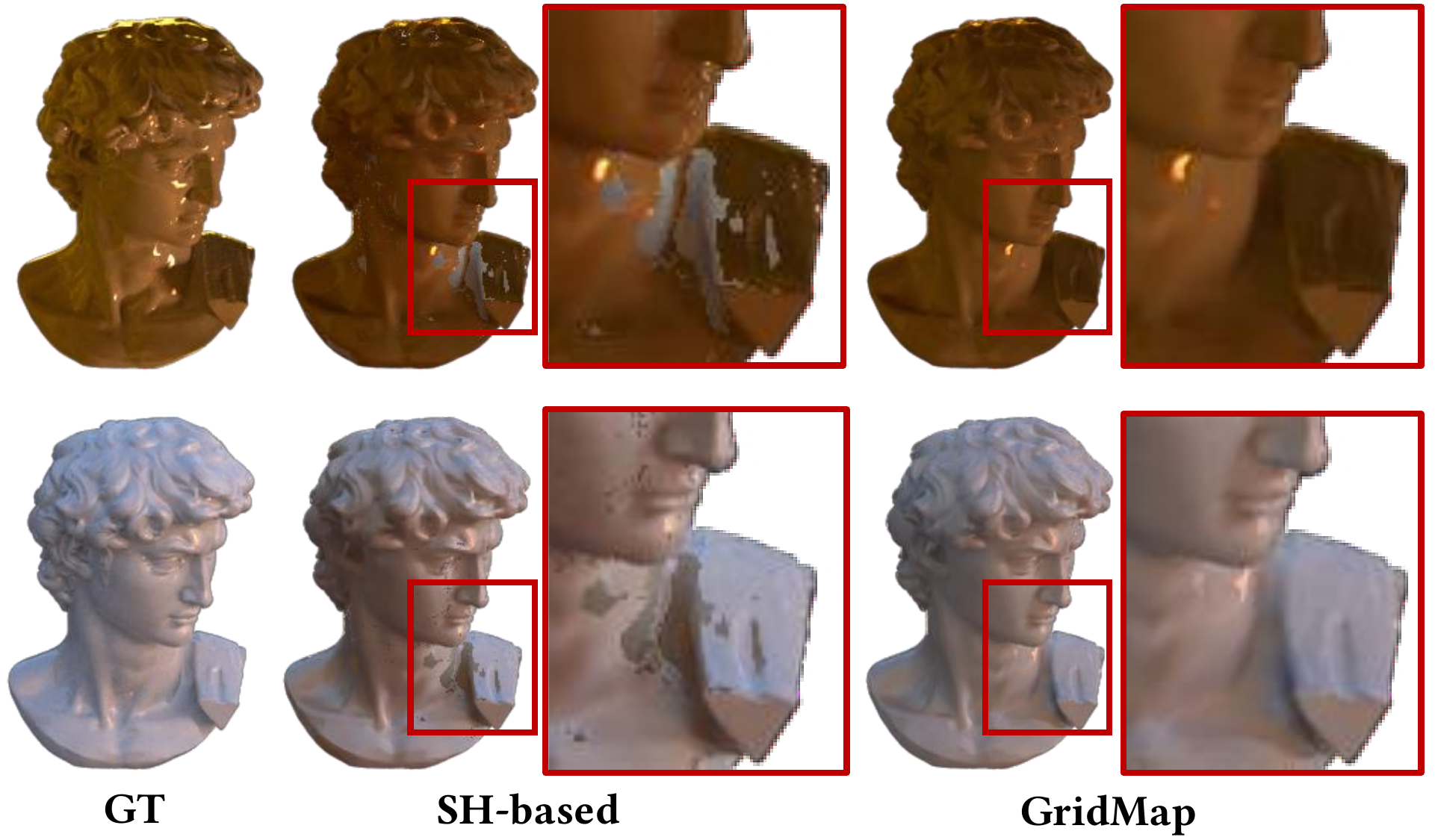}
    \caption{Relighting results using SH-based indrect light modeling and our GridMap.}
    \label{supp:sh-relight}
\end{figure}

\begin{table}
\centering
\caption{Chamfer Distance on the SMVP dataset. Optimal results are indicated by \textbf{bold }font.}
\label{supp:mesh_metric}
\vspace{-6pt}
\scalebox{0.85}{
\begin{tabular}{l|cccc} 
\hline
\multirow{2}{*}{Scene} & \multicolumn{4}{c}{Chamfer Distance $\downarrow$ (mm)}  \\ 
\cline{2-5}
& Ours  & PolGS & GIR    & GS-IR            \\ 
\hline
david                 & \textbf{6.535} & 6.537 & 62.075 & 77.032           \\
squirrel              & 9.527 & \textbf{6.604} & 13.118 & 28.148           \\
snail                 & \textbf{8.627} & 9.652 & 21.847 & 21.288           \\
\hline
\end{tabular}

}

\end{table}


\subsection{GridMap vs. SH-based Indirect Light}
In Figure~\ref{supp:sh-relight}, we compare between SH-based indirect light modeling (as utilized in Ref-Gaussian~\cite{yao2025refGS}) and our GridMap solution. We observe that the former leads to inconsistent artifacts during relighting, while our GridMap produces realistic transition from directly illuminated regions to those influenced by indirect light.

\footnotetext[1]{Training data for PANDORA and NeRSP models contain test viewpoints, and thereby their scores are only weakly comparable to others.}

\begin{figure}[t]
  \centering
  \includegraphics[width=0.99\linewidth]{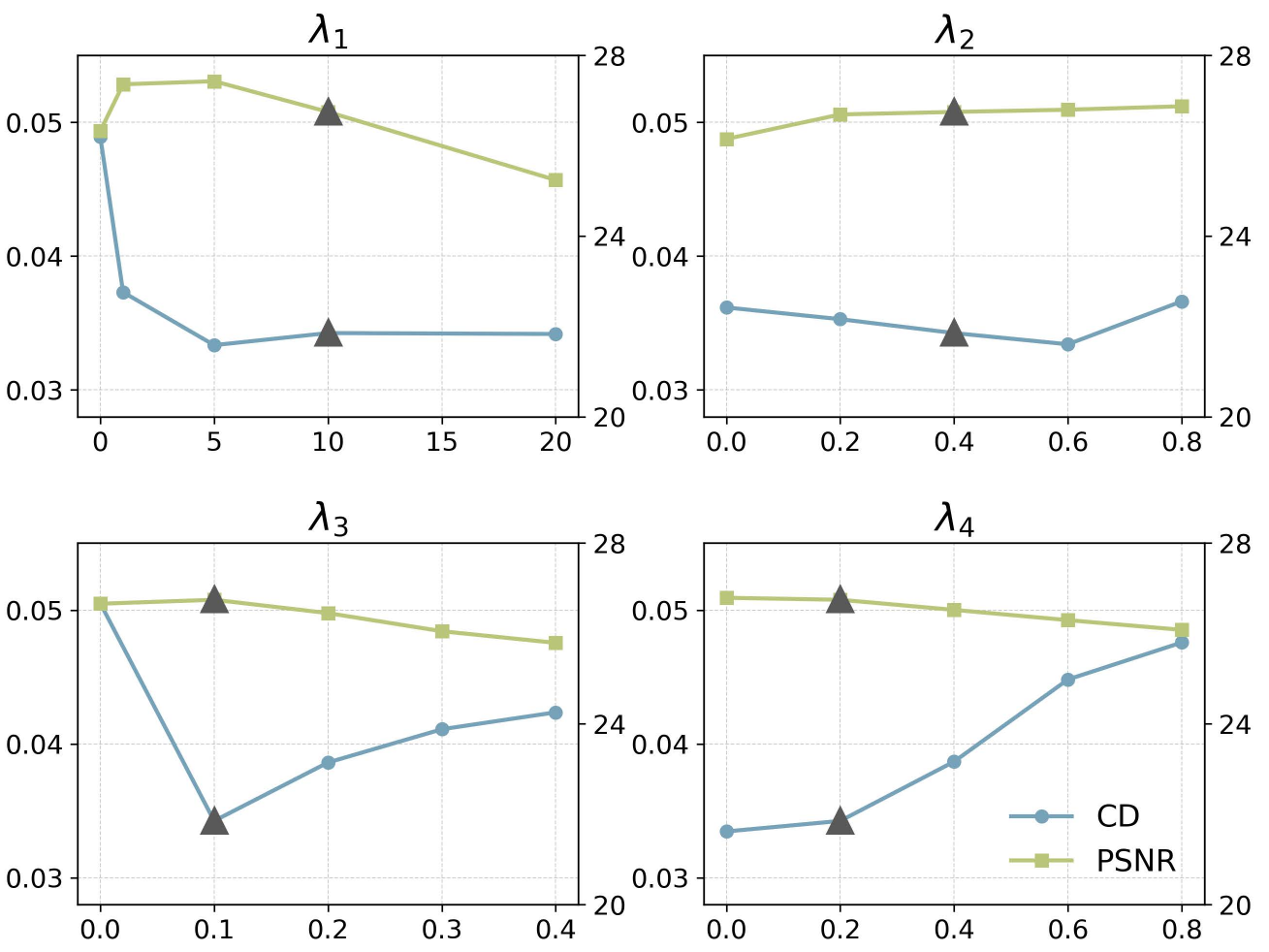}
   \caption{Cosine distance (CD$\downarrow$, left axis) and PSNR$\uparrow$ (right axis) across $\lambda$ choices. The reported values are indicated by triangles.}
   \label{fig:lambda}
\end{figure}

\subsection{Ablation on $\lambda$ Choices}

We conduct ablation study on different values of $\lambda$s to show that the reported values in the main paper are sufficiently close to optimality. See Figure~\ref{fig:lambda} for details.

\begin{figure}[t]
  \centering
  \includegraphics[width=0.99\linewidth]{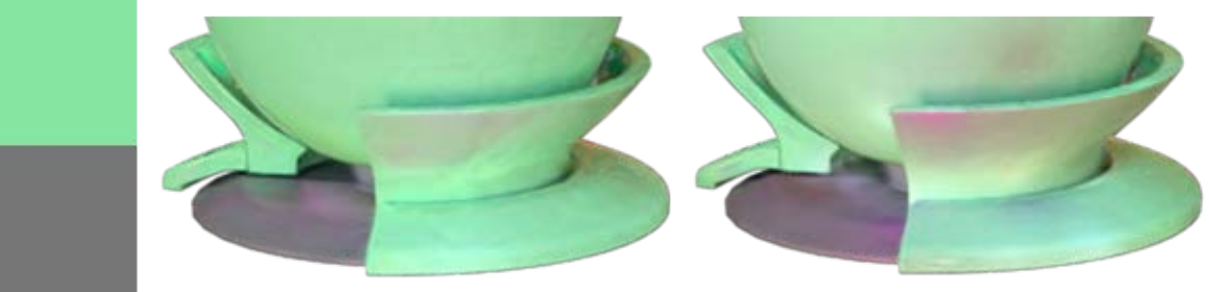}
   \caption{\textit{Left} ground-truth colors for object {\color{matgreen}body} and {\color{matgray}stand}. \textit{Mid} albedo reconstructed using GridMap. Object concavity and self-occlusion result in reddish artifacts. \textit{Right} albedo reconstructed using near-surface anchor placement. More severe color shifts are observed due to anchors being placed inside the object.}
   \label{fig:gridmap}
\end{figure}

\subsection{Failure Case and Comparison between Anchor Placement Strategies}

Our current strategy of placing anchor cameras on the bounding box of each object may fail under self-occlusion or objects with extreme shapes (\textit{e.g.} very thin or deeply concave). We observe this to be the main cause of failure during relighting in our experiments, and we have visually demonstrate it in Figure~\ref{fig:gridmap}.

Alternatively, we may choose to place anchors close to the object surface. However, this strategy also faces two issues: (a) Adaptively calculating anchor locations may introduce additional computation overhead during training. (b) Meshes constructed from TSDF do not fully align with Gaussian geometries, which can lead to anchors being sampled inside objects under this strategy and cause even larger errors. We compare two anchor placement strategies in Figure~\ref{fig:gridmap} to validate the latter claim.


\end{document}